%% file: access.tex
\documentclass{ieeeaccess}
\usepackage{cite}

\usepackage[hidelinks]{hyperref}
\usepackage[acronym,nomain,nonumberlist]{glossaries}
\usepackage{graphicx}%
\usepackage{multirow}%
\usepackage{amsmath,amssymb,amsfonts}%
\usepackage{amsthm}%
\usepackage{mathrsfs}%
\usepackage{textcomp}%
\usepackage{manyfoot}%
\usepackage{booktabs}%
\usepackage{algorithm}%
\usepackage{algpseudocode}%
\usepackage{listings}%
\usepackage{graphicx}
\usepackage{textcomp}
\usepackage{subcaption}
\usepackage{tabularx}

\usepackage{bm}
\makeatletter
\AtBeginDocument{\DeclareMathVersion{bold}
\SetSymbolFont{operators}{bold}{T1}{times}{b}{n}
\SetSymbolFont{NewLetters}{bold}{T1}{times}{b}{it}
\SetMathAlphabet{\mathrm}{bold}{T1}{times}{b}{n}
\SetMathAlphabet{\mathit}{bold}{T1}{times}{b}{it}
\SetMathAlphabet{\mathbf}{bold}{T1}{times}{b}{n}
\SetMathAlphabet{\mathtt}{bold}{OT1}{pcr}{b}{n}
\SetSymbolFont{symbols}{bold}{OMS}{cmsy}{b}{n}
\renewcommand\boldmath{\@nomath\boldmath\mathversion{bold}}}
\makeatother

\def\BibTeX{{\rm B\kern-.05em{\sc i\kern-.025em b}\kern-.08em
    T\kern-.1667em\lower.7ex\hbox{E}\kern-.125emX}}

\begin{document}
\include{acronym}

\history{Date of publication November 05, 2024, date of current version October 12, 2024.}
\doi{10.1109/ACCESS.2024.3492118}

\title{GraspLDM: Generative 6-DoF Grasp Synthesis using Latent Diffusion Models}
\author{\uppercase{Kuldeep R Barad}\authorrefmark{1}\authorrefmark{2}\IEEEmembership{Member, IEEE}, \uppercase{Andrej Orsula}\authorrefmark{1}\IEEEmembership{Member, IEEE},
\uppercase{Antoine Richard}\authorrefmark{1}\IEEEmembership{Member, IEEE}, \uppercase{Jan Dentler}\authorrefmark{2}, \uppercase{Miguel Olivares-Mendez}\authorrefmark{1}\IEEEmembership{Member, IEEE}, and \uppercase{Carol Martinez}\authorrefmark{1}\IEEEmembership{Member, IEEE}}

\address[1]{Space Robotics Research Group (SpaceR), Interdisciplinary Center for Security, Reliability and Trust (SnT), University of Luxembourg, 29 Ave. J. F. Kennedy, L-1855, Luxembourg}
\address[2]{Redwire Space Europe, 10 Rue Henri Schnadt, L-2530, Luxembourg}

\tfootnote{Code and resources are available at: \url{https://github.com/kuldeepbrd1/graspLDM}. \\ This work is supported by the Fonds National de la Recherche (FNR) Industrial Fellowship grant (15799985) and Redwire Space Europe.}

\markboth
{Barad \headeretal: Preparation of Papers for IEEE TRANSACTIONS and JOURNALS}
{Barad \headeretal: Preparation of Papers for IEEE TRANSACTIONS and JOURNALS}

\corresp{Corresponding author: Kuldeep R Barad (e-mail: kuldeep.barad@uni.lu).}

\begin{abstract}
Vision-based grasping of unknown objects in unstructured environments is a key challenge for autonomous robotic manipulation. A practical grasp synthesis system is required to generate a diverse set of 6-DoF grasps from which a task-relevant grasp can be executed. Although generative models are suitable for learning such complex data distributions, existing models have limitations in grasp quality, long training times, and a lack of flexibility for task-specific generation. In this work, we present GraspLDM- a modular generative framework for 6-DoF grasp synthesis that uses diffusion models as priors in the latent space of a \acrshort*{vae}. GraspLDM learns a generative model of object-centric $SE(3)$ grasp poses conditioned on point clouds. GraspLDM's architecture enables us to train task-specific models efficiently by only re-training a small denoising network in the low-dimensional latent space, as opposed to existing models that need expensive re-training. Our framework provides robust and scalable models on both full and single-view point clouds. GraspLDM models trained with simulation data transfer well to the real world without any further fine-tuning. Our models provide an 80\% success rate for 80 grasp attempts of diverse test objects across two real-world robotic setups.
\end{abstract}


\titlepgskip=-21pt

\maketitle

\section{Introduction}
\label{sec:intro}
Robotic grasping of unknown objects from visual observations is an integral skill for autonomous manipulation systems in the real world. Grasping is the fundamental manipulation task of generating restraining contacts on an object. An essential part of this task is grasp synthesis, which involves reasoning about a set of grasp configurations that are possible around an object. From this set of possible grasps, a subjectively good grasp can then be chosen based on grasp quality, task context, and kinematic feasibility. Real-world grasping requires a robot to determine grasping location and configuration from incomplete perceptual observations. Cameras are convenient as they are cost-effective and provide dense information in their field of view. However, the pixel-level information from a camera is unstructured and cannot be trivially mapped to geometric information that is helpful for manipulation, especially in unknown environments with object models that are not known apriori.

\begin{figure*}[ht]
\centering
    \includegraphics*[width=0.75\linewidth]
    {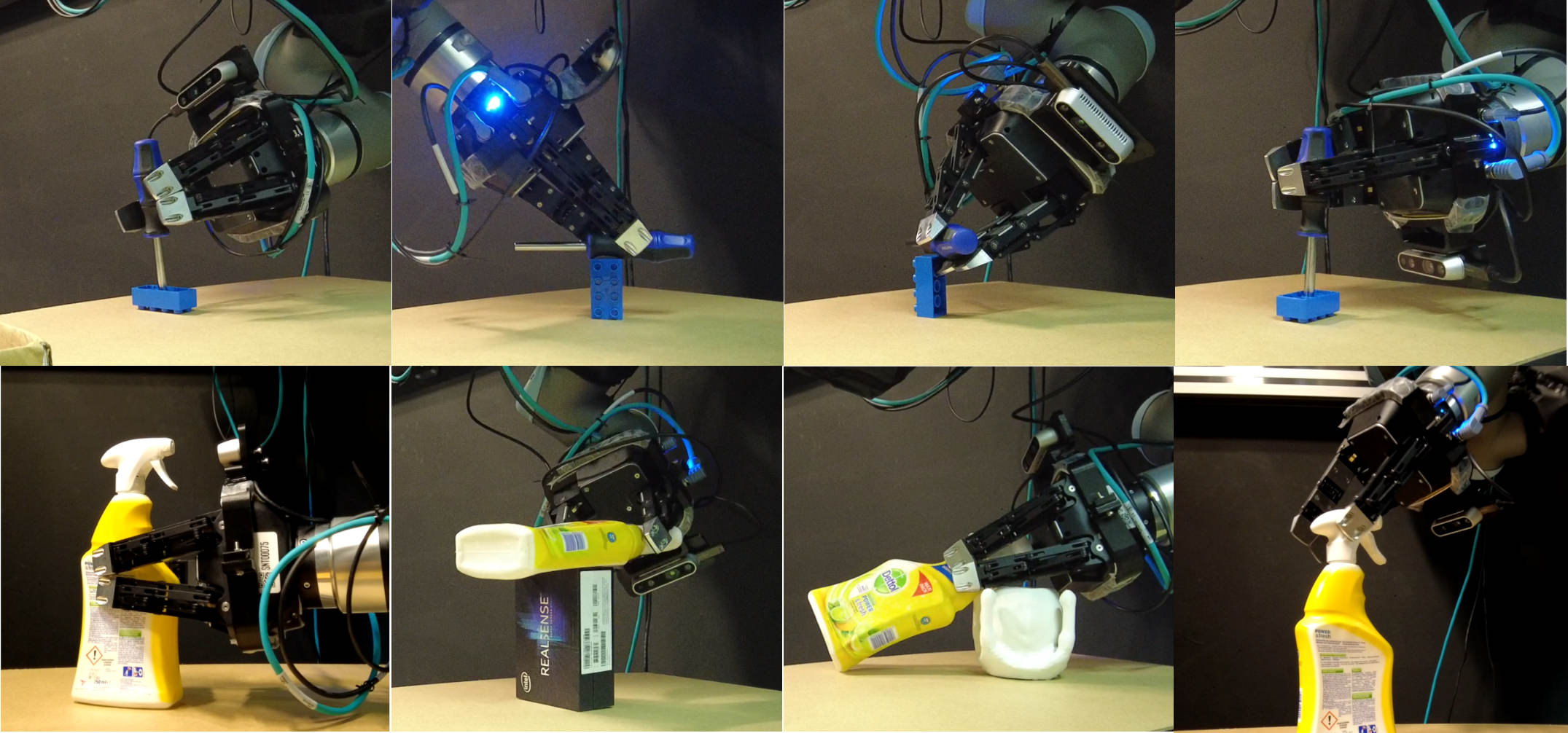}
    \caption{GraspLDM models trained on synthetic data successfully transfer to the real world and provide stable 6-DoF grasps from single-view RGB-D data in the presence of workspace and motion planning constraints.}
    \label{fig:graspldm_real_world}
\end{figure*}

For general-purpose robotic grasping, providing a single grasp configuration corresponding to an observation is not a suitable approach, as the grasp configuration might be kinematically unfeasible and contextually unsuitable for a task. On the other hand, we must develop systems that are task-agnostic but easy to adapt for specific tasks. This can be accomplished by designing vision-based grasp synthesis systems that more generally reason about a set of grasps around an object, further allowing selection based on a metric. However, vision-based grasp synthesis is a challenging problem for three main reasons. First, there are infinite possible grasps on any object and the distribution of good grasps can be complex, with multiple modes and discontinuities. Second, the grasp generation process has to rely on unstructured visual observations from the sensors. Finally, it needs to generalize to an arbitrary set of unknown objects that a robot may encounter in the real world. To alleviate some of these issues, contemporary works have increasingly relied on data-driven approaches~\cite{bohg2013data}. This is in contrast to analytical methods for grasp synthesis~\cite{bicchi2000robotic}, which assume the knowledge of an object's physical and geometric models and subsequently execute high-dimensional search satisfying an optimality criterion. Apart from the computational complexity, the grasps generated through such analytical grasp quality optimization could also transfer poorly to the real world~\cite{balasubramanian2012physical}.

Instead of the dynamics of hand-object contacts, data-driven grasp synthesis approaches focus on perceptual processing and generalizable representations~\cite{bohg2013data}. Early works focused on transferring grasps by using some notion of familiarity to a known reference~\cite{detry2012generalizing,el2008handling}. Subsequently, machine learning drove significant progress as the mapping from visual data to grasp configurations could be modeled using learned abstract representations. The Cornell dataset and the grasp rectangle representation introduced in~\cite{jiang2011efficient} stemmed a line of research~\cite{mahler2017dex,morrison2018closing} significantly advancing the state-of-the-art in grasping unknown objects. However, these approaches restrict the \acrfull*{dof} of grasping to a plane perpendicular to the image plane, typically used for top-down grasping from a table-top. In contrast, synthesizing 6-\acrshort*{dof} grasp configurations on unknown objects remains a challenge~\cite{newbury2022deep}.

Learning-based 6-\acrshort*{dof} grasp synthesis can be accomplished by supervised learning on annotated grasp data, reinforcement learning in relevant environments, or using exemplar learning by imitation. Reinforcement learning methods suffer from sample complexity and hyperparameter sensitivity~\cite{ibarz2021train}, while imitation learning is bottlenecked by human demonstration quality. In comparison, supervised learning has shown increased performance and generalization, with the availability of large and high-quality open-source datasets with dense grasp annotations~\cite{eppner2021acronym,fang2020graspnet}. Within supervised learning, however, most approaches are based on direct regression of grasp poses or grasp quality. Fewer works~\cite{mousavian20196,murali20206} employ generative modeling. Generative modeling approaches learn the structure of the input data and efficiently model the data-generating distribution. Generative models~\cite{kingma2013auto,ho2020denoising,song2019generative} have consistently improved performance in real-world generation and synthesis across domains like natural language and computer vision. In particular, a new class of generative models called \acrfull*{ddm} have recently shown state-of-the-art performance on likelihood-based metrics and output sample quality~\cite{sohl2015deep,ho2020denoising,song2020score}. Given the successful application of \acrshort*{ddm}s to complex generation tasks for images~\cite{ho2020denoising} and point clouds~\cite{zeng2022lion}, we investigate their potential for improving generative modeling in robotics.





We focus on the problem of generating object-centric 6-\acrshort*{dof} grasps using point cloud observations. We assume a parallel jaw gripper and parameterize a grasp by its $SE(3)$ pose. The objective is to learn a representation of object-centric 6-DoF grasps that can generalize to unknown objects. Learning this representation is difficult, as the model must reason about an arbitrary set of grasp poses for each object in the $SE(3)$ space. Deep generative models are suitable for this problem, as they are designed to learn complex data-generating distributions efficiently. 6DoF-Graspnet~\cite{mousavian20196}, an early generative grasp synthesis model, learns the distribution of object-centric grasps in a continuous latent space of a conditional \acrfull*{vae}. However, the \acrshort*{vae} exhibits poor sampling quality and requires an additional iterative post-processing stage to improve grasp samples. The shortfalls of the \acrshort*{vae} to learn effective representations could be attributed to the \textit{prior gap} problem in \acrshort*{vae}s~\cite{zhao2017infovae}. The \textit{prior gap} problem refers to the gap between the posterior distribution modeled by the \acrshort*{vae} encoder and the assumed prior, which results in poor quality samples when decoding random latents from the prior. On the other hand, diffusion models~\cite{song2019generative, ho2020denoising} can overcome these problems and are simpler to train. In particular, the flexible generation architecture of latent diffusion~\cite{rombach2022high} is especially appealing.

\begin{figure}[t]
    \centering
    \includegraphics[width=\linewidth]{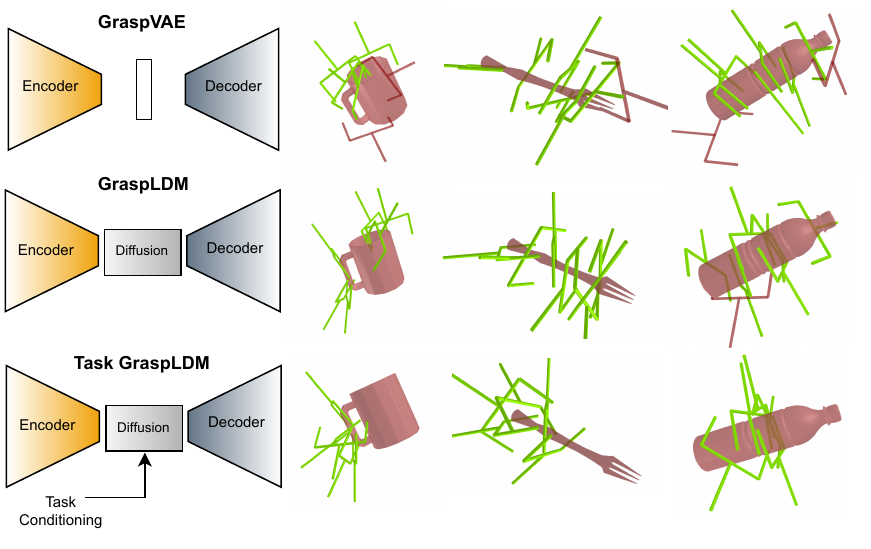}
    \caption{GraspLDM uses a denoising diffusion model in the latent space of a \acrshort*{vae} to improve grasp generation performance. It also enables injection of task-conditional guidance in a modular manner.}
    \label{fig:intro-illustration}
\end{figure}

In this work, we propose GraspLDM - a learning-based generative modeling approach for 6-DoF grasp synthesis. GraspLDM learns the distribution of successful grasps on object point clouds using latent diffusion that transfers to the real world (Fig.~\ref{fig:graspldm_real_world}). In GraspLDM, a diffusion model is trained efficiently inside the low-dimensional latent space of a VAE. This diffusion model acts as a prior and bridges the gap between the \acrshort*{vae} prior and the posterior distribution fit during the training. Using diffusion in the latent space improves the quality of sampled grasps and avoids the representational complexity of operating directly on pairs of point clouds and grasp pose. It especially enables two aspects of model flexibility that are not available in existing generative models for grasp synthesis using \acrshort*{vae}~\cite{mousavian20196} and diffusion~\cite{urain2022se} alone. First, task-specific conditional guidance can be provided solely in the latent space without re-training the \acrshort*{vae} encoder and decoder. Second, multiple diffusion models can be trained efficiently and plugged inside this low-dimensional latent space. Further, the choice of latent diffusion structure is justified by the high dimensionality of representation required to directly operate on a pair of point cloud and grasp in a vanilla diffusion model.  

In summary, our work makes three contributions. \textbf{(1) We introduce a new generative modeling framework for \mbox{6-DoF} grasp synthesis using latent diffusion.} To the best of our knowledge, no other work has applied latent diffusion for 6-DoF grasp synthesis for scalable real-world parallel-jaw grasping. \textbf{(2) We show that a diffusion model in the latent space improves the grasp sample quality of a standard \acrshort*{vae} model.} In simulation tests, our latent diffusion models improve generation performance. Further, they transfer to the real world to provide more stable grasps from single-view point clouds. \textbf{(3) We demonstrate that our architecture enables the injection of task-specific conditioning in generation with limited additional training effort.} The separation of \acrshort*{vae} and diffusion model introduces flexibility that allows rapid training of task-specific models in the latent space.



\section{Related Work}
\subsection{6-DoF Grasp Synthesis}
Grasp synthesis was initially studied using analytical methods~\cite{bicchi2000robotic}, which rely on the knowledge of object models and properties that are rarely available in the real world. The complexity of the search problem and the dependence on models paved the way for methods that learn directly from data. To learn from data, the focus changed from contact modeling to designing an effective visual processing system and choosing robust representations. Such works initially relied on primitive decomposition~\cite{el2008handling}, demonstrations~\cite{ekvall2007learning}, template matching~\cite{ciocarlie2014towards} or transferring some notion of grasp familiarity~\cite{detry2012generalizing}. More recently, grasp synthesis has been framed as the problem of finding stable grasp poses directly from sensor data, subject to a gripper. Data-driven and learning-based methods have made notable progress, especially on 4-DoF planar grasp synthesis using parallel-jaw and vacuum grippers. In contrast, generalizable 6-DoF grasp synthesis is an active research topic~\cite{newbury2022deep} with two broad classes: discriminative and generative approaches. Discriminative approaches rely on a manual pose sampling strategy and learn to discriminate good grasps from bad grasps using a quality measure~\cite{mahler2017dex}. Generative approaches instead directly learn to generate grasp poses from an observation~\cite{sundermeyer2021contact, mousavian20196} with implicit sampling. Since grasps can be sampled directly from the learned model~\cite{mousavian20196} these tend to be efficient. Some works on 6-DoF grasping also address auxiliary tasks to improve grasp generation, like shape completion and 3D reconstruction~\cite{jiang2021synergies}. In this work, we present a generative approach that learns a continuous distribution of object-centric grasps conditioned on point clouds in an end-to-end manner.

\subsection{Generative Models for Grasp Synthesis}
A conditional \acrshort*{vae}~\cite{kingma2013auto} was introduced in~\cite{mousavian20196} to learn a point cloud conditioned distribution of good grasps in a continuous latent space, from which they can be efficiently sampled. While the \acrshort*{vae} is successfully able to cover multiple grasp modes, \acrshort*{vae} alone provides a large fraction of unsuccessful grasps when executed. Consequently, additional stages for pose refinement are required.~\cite{murali20206} builds upon~\cite{mousavian20196} with an additional stage of collision checking using a learned network. However, VAEs underperform against discrete regression models~\cite{sundermeyer2021contact}. GraspLDM improves the generative performance over~\acrshort*{vae}-based methods using latent diffusion. 

\subsection{Denoising Diffusion Models}

GraspLDM utilizes a diffusion model based on \acrfull*{ddpm}\cite{ho2020denoising}. \acrfull*{sgm}~\cite{song2019generative} and \acrfull*{sde} are equivalent formulations of the diffusion process with notable implementation differences. For grasp pose generation,~\cite{urain2022se} uses the \acrshort*{sgm} formulation and learns a scalar field representing the denoising score. While their models effectively refine randomly sampled SE(3) grasp poses to low-cost (good grasp) regions, the analysis only presents learning on full point clouds of a single category (Mugs). Real-world tests on Mugs in~\cite{urain2022se} use the ground truth pose of the object. Further, Langevin-type sampling is slow and the model needs to be retrained for downstream tasks. In contrast, GraspLDM is a more flexible formulation to~\cite{urain2022se} that operates a diffusion model on a vector field in the continuous latent space of a \acrshort*{vae}. Our analyses and validation are also more extensive using full and partial point clouds from a large object category set, complemented with real-world transfer tests. GraspLDM can also use fast samplers like \acrfull*{ddim} as a drop-in replacement post-hoc with minimal performance loss. Further, GraspLDM can be extended to gripper parameterizations beyond the pose, in contrast to~\cite{urain2022se}. Diffusion models have also been utilized for dexterous grasp synthesis~\cite{li2022gendexgrasp}, but rely on the availability of the object model.   


\begin{figure*}[t]
    \centering
    \includegraphics[width=\textwidth]{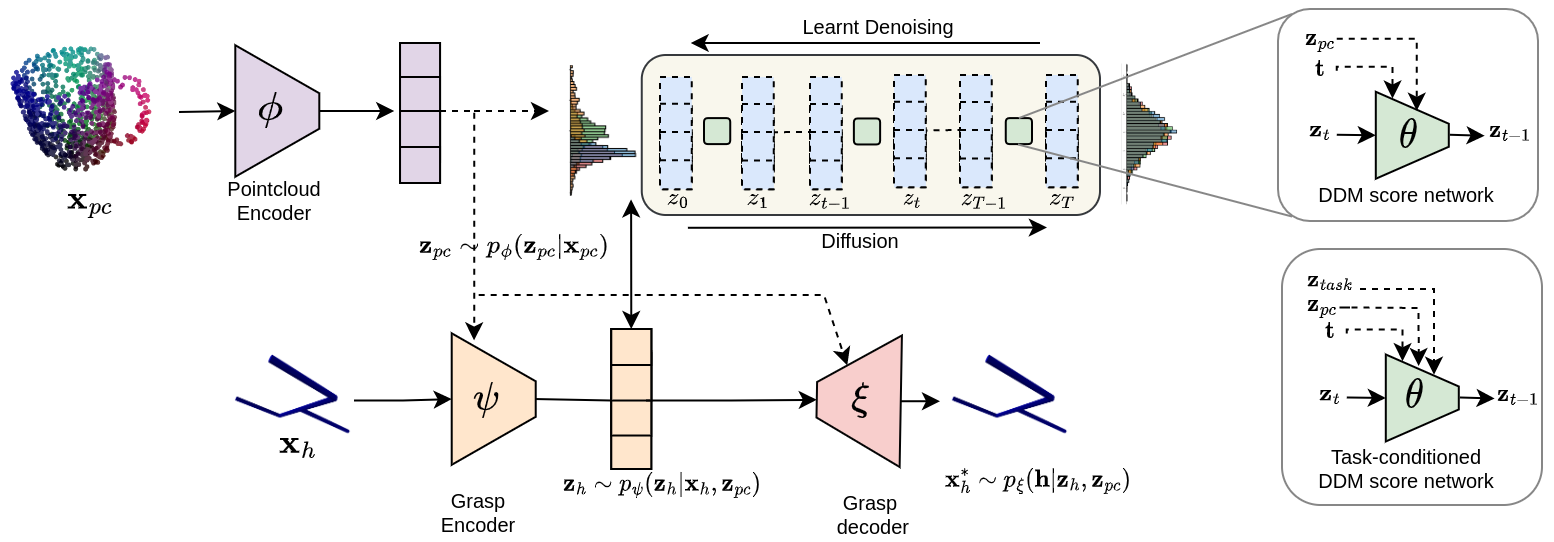}
    \caption{Grasp Latent Diffusion Model (GraspLDM) is composed of a point cloud encoder ($\phi$), a grasp encoder ($\psi$), a grasp decoder ($\xi$), and a latent diffusion module using a score network ($\theta$). The point cloud encoder encodes a point cloud into a shape latent ($\mathbf{z}_{pc}$). At test time, the grasp encoder is not required and we sample the grasp latent $\mathbf{z}_h$ directly from the prior distribution. This latent goes through reverse diffusion before decoding. For task conditional generation, we modify the diffusion score network to accept task context $\mathbf{z}_{task}$.}
    \label{fig:grasp_ldm_arch}
\end{figure*}
\section{Grasp Latent Diffusion Models}

We consider the problem of generating 6-DoF grasps on unknown object point clouds. We cast it as a generative modeling problem that is concerned with learning the conditional distribution $p(H \,| \,\mathbf{x_{pc}})$, where ${H} \in \mathit{SE(3)}$ is a successful grasp pose, given a point cloud $\mathbf{x_{pc}} \in \mathbb{R}^{3\text{x}n}$. Learning effective generalizable representations to model this distribution is difficult because of its complexity in $SE(3)$ space. To tackle this, we propose to learn a latent variable model $p(H | \mathbf{z}, \mathbf{x_{pc}})$, where $\mathbf{z}$ represents the latent, by maximizing the likelihood of the training data consisting of successful grasps.

\begin{equation}
    p(H| \mathbf{x_{pc}}) \, = \, \int \, p(H| \mathbf{z}, \mathbf{x_{pc}}) p(\mathbf{z}) d\mathbf{z}
    \label{eq:likelihood}
\end{equation}

Since we do not have access to the real latent space, computing the likelihood of data over the whole latent space in Eq.~\ref{eq:likelihood} is intractable. To overcome this, we use a \acrshort*{vae} formulation for learning our generative model. VAE's encoder compresses the point cloud-grasp pair input to a continuous latent space with a prior distribution. However, the approximate posterior of the encoder does not match the prior perfectly, especially if it prioritizes data reconstruction for grasp accuracy. To tackle this, we use a diffusion model in its latent space. Together, they form the latent diffusion model for grasp generation that we call GraspLDM. The diffusion model simultaneously improves sample quality by bridging the prior gap in the latent space and enables the introduction of latent space guidance with task context. The following sections describe the GraspLDM design in detail.

\subsection{Conditional Variational Autoencoder}
Our conditional \acrshort*{vae} structure consists of a point cloud encoder ($\phi$), a grasp pose encoder ($\psi$), and a grasp pose decoder ($\xi$) as shown in Fig.~\ref{fig:grasp_ldm_arch}. We use the point cloud encoder $q_{\phi}(\mathbf{z_{pc}|\mathbf{x}_{pc}}) : \mathbf{x}_{pc}\in\mathbb{R}^{3\times n} \mapsto \mathbb{R}^{m}$ that can operate on unordered point-sets of size $n\in\mathbb{N}^+$ to provide a fixed size latent ($\mathbf{z}_{pc}$) of size $m\in\mathbb{N}^+$ called the shape latent. The shape latent is used as conditioning in the grasp pose encoder $p_{\psi}(\mathbf{z}_h | H, \mathbf{z}_{pc})$ and the grasp pose decoder $p_{\xi}(H| \mathbf{z}_h, \mathbf{z_{pc}})$. $\mathbf{z}_h$ is the conditional grasp latent at the VAE bottleneck. Finally, the model is trained by jointly optimizing the parameters ($\psi, \theta, \xi$) of the encoders and decoders to maximize the Evidence Lower Bound (ELBO)~\cite{kingma2013auto}:
\begin{align}
   \mathcal{L}_{ELBO} (\phi, &\psi,  \xi) \, =  \, \mathbb{E} 
   \bigl[ \; \log\, p_{\xi}(H^{*}| \mathbf{z}_h, \mathbf{z}_{pc}) \, \label{eq:elbo}  \\ 
   & - \lambda \,D_{KL} \left(q_{\psi}(\mathbf{z}_h | H, \mathbf{z}_{pc}) \, || \, \mathcal{N}(\mathbf{0,I}) \right) \; \bigr] \nonumber
\end{align}
In Eq.~\ref{eq:elbo}, the first term is the likelihood of the decoder reconstructing the input data. The second term measures the divergence between the approximate posterior distribution $q_{\psi}(\mathbf{z}_h | H, \mathbf{x}_{pc})$ of the encoder and the assumed prior, $\mathbf{z}\sim\mathcal{N}(0,1)$. $\lambda$ is the weighting hyperparameter that controls the strength of KL regularization in the latent space. A constant $\lambda=1$ provides the strict \acrshort*{elbo} objective. However, this frequently results in the KL divergence term ($D_{KL}$) becoming vanishingly small early in the training. When this happens, the conditional decoder becomes auto-regressive and the latent has no effect on the output~\cite{zhao2017infovae}. To avoid this, we use linear annealing on the $\lambda$ parameter. 

\subsection{Latent Diffusion}
In principle, sampling from \acrshort*{vae}'s prior and decoding it should be sufficient to generate successful grasps. However, decoding a latent drawn from the prior often results in lower sample quality because of the prior gap problem introduced in Section~\ref{sec:intro}. To alleviate this issue, we propose to use a \acrfull*{ddm} in the latent space. We justify the choice of latent diffusion structure from two observations. First, a simple \acrshort*{ddpm} would need to encode a point cloud at each step and use a high-dimensional intermediate representation, increasing computation for each denoising step. Alternatively, separating the point cloud encoder from the diffusion requires separate expensive training of a point cloud auto-encoder. Further, such a model needs to be re-trained for newer task conditionings. 

A \acrshort*{ddm} consists of two processes: a forward diffusion process and a reverse denoising process. In this work, we use the standard discrete-time DDPM~\cite{ho2020denoising}, where the forward process is a linear Markov chain of deterministic Gaussian kernels in Eq.~\ref{eq:gaussian_diff_kernel}. The noise added at each time-step $t \in [1,T]$ is pre-defined by a variance schedule $\beta_t$. We use linearly increasing $\beta_t$ such that the distribution $q(\mathbf{z}_T)$ converges to the standard normal distribution at the final forward time step $T$. In the forward diffusion process, a noisy sample at any time step can then be obtained by Eq.~\ref{eq:diffusion_fwd_sampling}.
\begin{subequations}
\begin{align}
    q(\mathbf{z}_t | \mathbf{z}_{t-1}) \; = \; \mathcal{N}(\sqrt{1-\beta_t} \mathbf{z}_{t-1}, \beta_t \mathbf{I}) 
    \label{eq:gaussian_diff_kernel} \\ \vspace{2mm}
    \mathbf{z}_t \, = \, \sqrt{\Bar{\alpha}_t} \, \mathbf{z}_0 \, + \, (1 - \Bar{\alpha}_t)\bm{\epsilon}\; ; \; \;\;\;\; \label{eq:diffusion_fwd_sampling} \\ 
    \Bar{\alpha}_t \; = \; \prod_{t=1}^T \, ( 1- \beta_t) \quad ; \quad \bm{\epsilon} \sim \mathcal{N}(\mathbf{0,I}) \nonumber
\end{align}

\end{subequations}
Intuitively, the forward diffusion process adds noise to input data until the data distribution transitions to a simple prior distribution. Consequently, recovering the data distribution from the prior requires a time reversal of the forward process from $t=T$ to $t=0$. This reverse diffusion process uses a learned score network ($\theta$) which de-noises a sample, in our case the latent $\mathbf{z}_h$, from $t$ to $t-1$. For notational simplicity, we use $\mathbf{z}_t$ to refer to conditional grasp latent $\mathbf{z}_h$ at time $t$.  Using the reverse diffusion kernel in Eq.~\ref{eq:reverse_diff_kernel}, the distribution of the de-noised sample at time-step $t-1$ can be modeled by the mean $\mu_{\theta}(z_t, t)$ and a known variance $\sigma_t^2$. More simply, we can re-parameterize the mean and instead learn a network to directly predict the noise $\bm{\epsilon}_{\theta}(\mathbf{z}_t, t)$ to be removed at each step~\cite{ho2020denoising}. The denoising score model $\bm{\epsilon}_{\theta}$ predicts the noise to be removed, conditioned on the current time-step, while the parameter weights ($\theta$) are shared across all the time steps. Then, the sampling can be done using Eq.~\ref{eq:ancestral_sampling}, where $\sigma_t$ can be simplified to use variance similar to the forward process i.e. $\beta_t$. DDMs with this formulation can be trained to maximize an ELBO-like objective. Given the simplified formulation in~\cite{ho2020denoising} and adapting it to our latent diffusion architecture, this objective $\mathcal{L}_{D} (\theta)$ can be written as Eq.~\ref{eq:loss_diffusion}.

\begin{equation}
    p_{\theta}(\mathbf{z}_{t-1}|\mathbf{z}_t) \, = \, \mathcal{N}(\mu_{\theta}(z_t, t), \sigma_t^2\mathbf{I})
    \label{eq:reverse_diff_kernel}
\end{equation}
\begin{equation}
    \mathbf{z}_{t-1} = \frac{1}{\sqrt(1 - \beta_t)} \bigl( \mathbf{z}_t - \frac{\beta_t}{1-\Bar{\alpha_t}} \bm{\epsilon}_{\theta}(\mathbf{z}_t, t) + \sigma_t \eta \bigr)
    \label{eq:ancestral_sampling}
\end{equation}

\begin{align}
    \mathcal{L}_{D} (\theta) \, = & \, \mathbb{E} 
     || \bm{\epsilon}_t - \bm{\epsilon}_{\theta}(\mathbf{z}_{h,t},\mathbf{z}_{h,0}, t) ||^2
     \label{eq:loss_diffusion}
\end{align}

\subsection{Implementation Details}
The training is done in two stages. In the first stage, we fit the parameters of encoders and decoders by maximizing the VAE ELBO in Eq.~\ref{eq:elbo}. In the second stage, we fit the parameters of the score model to minimize DDM loss in Eq.~\ref{eq:loss_diffusion}. While it is possible to train all the networks in a single stage, it is easier and faster to train the diffusion model once the latent space of the VAE is frozen. In single-stage training, the constantly changing posterior distribution at the encoder means that the optimization of the weights of the denoising score network is wasteful until VAE training has converged sufficiently. To train our networks, we use objects and grasp annotations from the ACRONYM~\cite{eppner2021acronym} dataset. We use train-test splits defined in~\cite{sundermeyer2021contact}. We empirically choose the \acrshort*{elbo} modifying hyperparameter $\lambda$, which is annealed linearly from $1e-7$ to $0.1$ for $50\%$ of the training steps and then held constant until the end of training. This ensures that the KL term does not vanish in the early stages of the training while also enforcing lower regularization to prioritize grasp pose reconstruction during the later stages. 

The point cloud encoder is based on Point-Voxel CNN (PVCNN) architecture~\cite{liu2019point} which is more efficient than conventionally used PointNet architecture~\cite{liu2019point}. The input to the point cloud encoder is fixed at 1024 points, to balance memory usage with point sampling density. The output of the point cloud encoder is a latent ($\mathbf{z}_{pc}$) of size 128. The grasp pose encoder and decoder are 1D convolutional neural networks with residual blocks. Conditioning is done using FiLM~\cite{perez2018film}, which uniquely scales and shifts the feature maps at every residual block based on the conditioning features. The decoder network is the same as the encoder but reversed, whose output is the reconstruction ($\mathbf{x}^*_h$) of the input. The noise prediction network for diffusion is a residual network similar to the grasp encoder and is conditioned on both the diffusion time step ($t$) and the shape latent ($\mathbf{z}_{pc}$). For this purpose, we first embed the time step scalar to the desired dimensions using sinusoidal positional embedding. The shape latent is then projected onto the same dimension and added to the time embedding, before applying it to the features at every block.
\begin{align}
    \mathbf{h} = \begin{bmatrix}
        \mathbf{t}, \mathbf{a} 
    \end{bmatrix}^T  ; \quad \mathbf{a} = \frac{4}{1+q_w} \; 
    \begin{bmatrix}
      q_x, q_y, q_z 
    \end{bmatrix} ^T
    \label{eqn:define_mrp}
\end{align}

We use the representation of the grasp pose ($\mathbf{h}$) given in Eq.~\ref{eqn:define_mrp}, where $\mathbf{t}$ is the translation vector while $\mathbf{a}$ is the \acrfull*{mrp}~\cite{crassidis1996attitude} for orientation. Eq.~\ref{eqn:define_mrp} also relates \acrshort*{mrp} with quaternions where $q_w$ is the quaternion scalar component. In contrast to regressing pose matrices or quaternion vectors that have one or more dependent parameters, the three parameters of \acrshort*{mrp}s can be regressed independently.

We train our models on both full and partial point clouds. In each case, the point clouds are augmented online with random rotations, point jitter, and point dropout. Random rotation is sampled by selecting a random unit vector (axis) and a random magnitude (angle). Point jitter randomly perturbs the 3d coordinate of the points with a standard deviation of 1cm in all directions around the original to emulate jitter artefacts of a noisy sensor stream. Point dropout randomly reduces the availability of unique points by removing up to 40\% of the input points. Since the input point cloud has to be of a fixed shape, the removed points are replaced by duplicating existing points. To avoid relying on an explicit canonical frame, the point cloud and grasp poses are both expressed in a frame with the origin at the centroid of the input point cloud. 

For the first term of the \acrshort*{elbo} (Eq.~\ref{eq:elbo}) loss that measures reconstruction, we use L2 loss between inputs and outputs of the 6-parameter grasp pose. For learning rate decay, we scale down the learning by 0.1 every 1/3rd of the total training steps, going from 1e-3 to 1e-5 to avoid large optimization steps in later stages of the training. For the diffusion model, the noise variance ($\beta_t$) for forward diffusion is set to a linear schedule $\beta$ with $\beta_0 = 5e-5$ and $\beta_T = 1e-3$. The denoising diffusion formulation closely follows the original DDPM formulation with 1000 time steps in the nominal configuration.

\section{Results and Discussion}

We conduct a thorough evaluation of our models in four segments. First, the 6-DoF generation capability of our models is evaluated on full point clouds sampled from object meshes in Section~\ref{sec:full-pc-generation}. We then investigate the benefits of using latent diffusion by evaluating class-conditioned generation in Section~\ref{sec:results-condgen} and the flexibility of using a drop-in reverse diffusion sampler for speed improvements in Sec~\ref{sec:results-reverse-diff-sampler}. In Section~\ref{sec:results-partial-pc}, we evaluate learning on partial point clouds from a single RGB-D image. Finally, we validate the practical effectiveness of our models and sim-to-real transfer with two real-world robotic setups in Section~\ref{sec:results-real-world}.  

\label{sec:results}
\subsection*{Evaluation}

\begin{figure}[t]
    \centering\includegraphics[width=\linewidth]{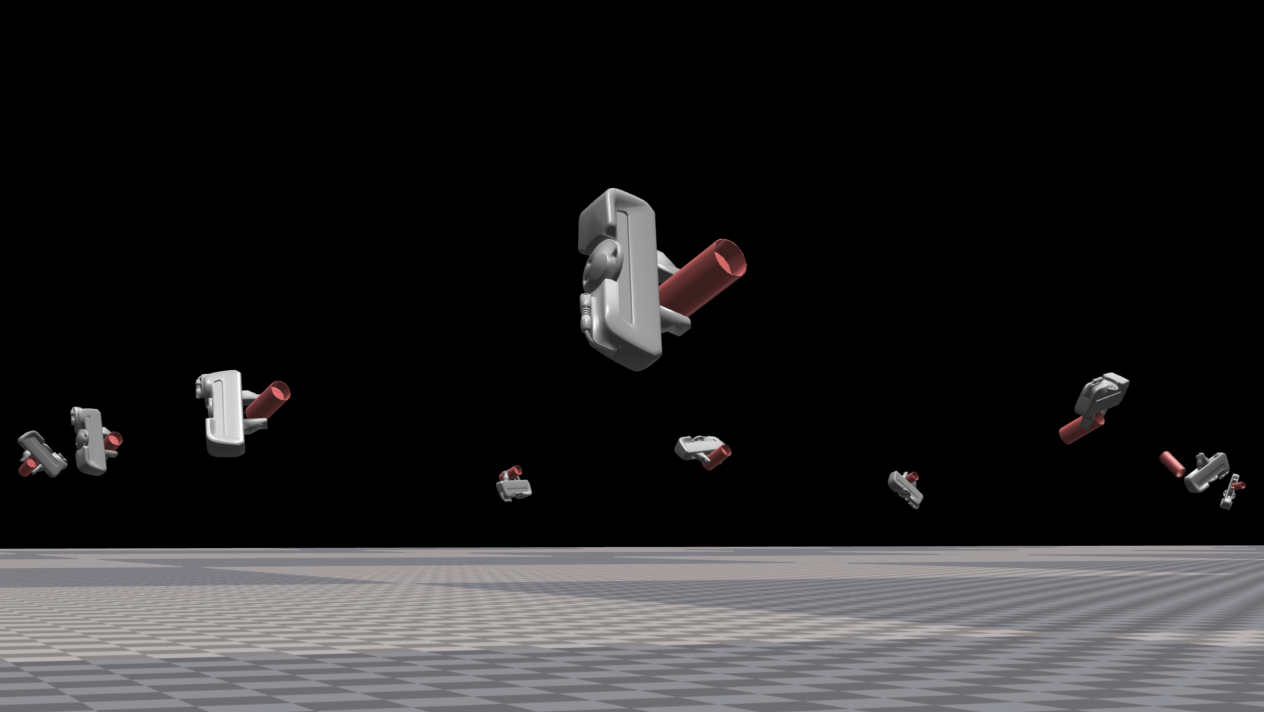}
        \caption{Multi-object grasping environments with Franka gripper in Isaac Gym for success rate evaluation.}
        \label{fig:sim}
\end{figure}

\begin{figure*}[t]
    \centering
    \begin{subfigure}[t]{0.5\textwidth}
        \centering
        \includegraphics[width=0.95\linewidth]{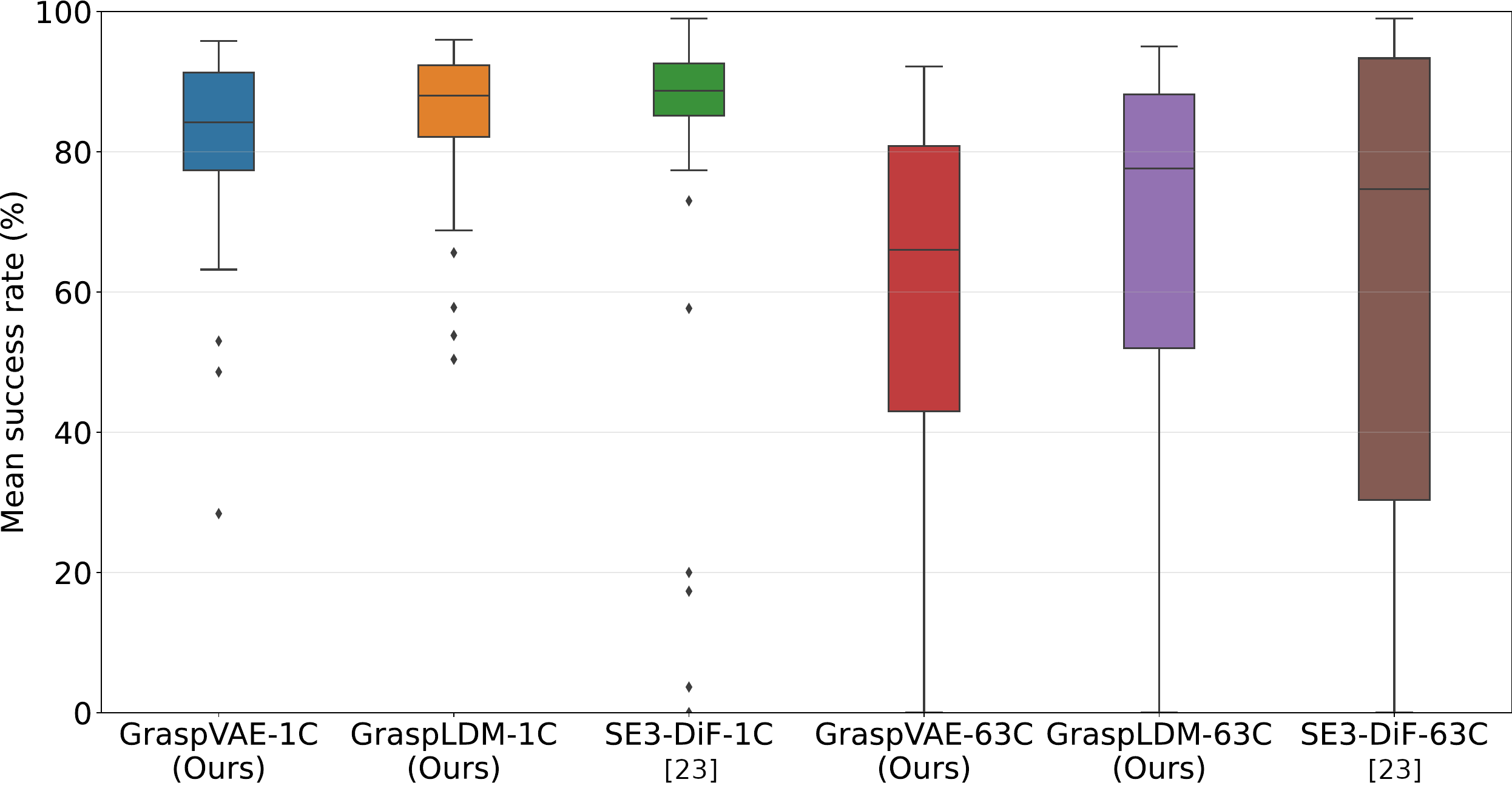}
        \caption{}
        \label{fig:full_pc_success_rate}
    \end{subfigure}%
    \begin{subfigure}[t]{0.5\textwidth}
        \centering
        \includegraphics[width=0.95\linewidth]{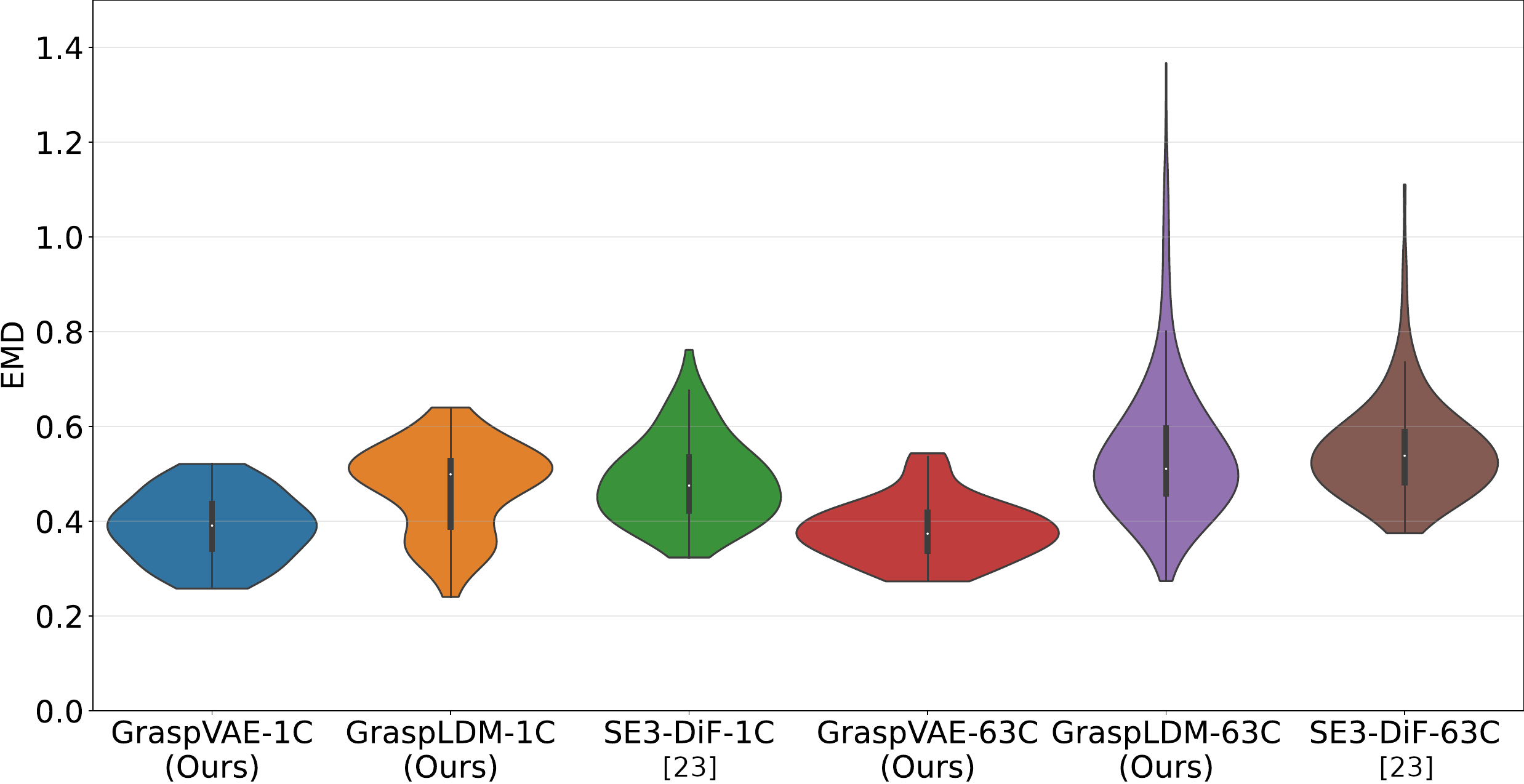}
        \caption{}
        \label{fig:full_pc_emd}
    \end{subfigure}
     
\caption{Grasp generation performance and scaling on full object point clouds ($N=1024$). (a) The mean \textit{success rate} ina  simulation of 300 generated grasps poses per object. (b) SE(3) \acrshort*{emd} between ground-truth grasp pose distribution and 100 sampled grasp poses (lower is better). SE3-DiF-1C and SE3-DiF-63C are the SE(3) Grasp Diffusion models from~\cite{urain2022se}.}
\end{figure*}

We evaluate the performance and scaling of GraspLDM on two category sets- 1C and 63C, where C denotes the number of ShapeNetSem categories used from the ACRONYM dataset. 1C is composed of 110 Mugs in the train set and 50 held-out Mugs in the test set. The 63C set contains 1100 objects from 63 categories in the train set and 400 held-out objects in the test set. In the case of both full and partial point clouds, the point cloud is randomly rotated, noised, and regularized to 1024 points during training and testing before it is fed to the model.

We primarily evaluate the \textit{success rate} of the grasp poses generated by the models. In Section~\ref{sec:full-pc-generation} \&~\ref{sec:results-partial-pc}, we assess the success rate with a large-scale parallel simulation in Isaac Gym~\cite{makoviychuk2021isaac}. Simulation offers the best method of exhaustively evaluating 6-DoF grasp generation. Real-world testing of 6-DoF grasps is limited by factors like kinematic feasibility, external collisions, and the time required per grasp execution. Therefore, we first evaluate the generator models in the simulation and subsequently validate the performance transfer to the real world. In the simulation, we use a two-fingered gripper (Franka hand) without the robotic arm as shown in Fig.~\ref{fig:sim} and ignore gravity to eliminate the effect of external factors on the evaluations. The simulation executes three steps for each grasp pose to report a grasp's success or failure. (1) The gripper is spawned at the given grasp pose relative to the object with the fingers open. (2) The fingers are closed at the grasp pose. (3) The gripper shakes the object in X and Y directions and then lifts the object 1m in the +Z direction. The grasp is successful if the object remains attached to the gripper. Note that the evaluation is conservative as any penetration of the gripper in the first step or adverse contact in the second stage will result in a grasp failure. We observe that for large objects, ground truth grasp poses in the ACRONYM dataset provide a very low or zero success rate in our simulation. Therefore,  we filter out the objects in each set for which the ground truth grasps fail more than 75\% of the time. 

While the success rate evaluates the quality of the generated grasps, a model with a high success rate may generate grasps only around a small region of the object, instead of covering all the graspable regions around the object. Consequently, we also use SE(3) \textit{\acrfull*{emd}} metric~\cite{urain2022se} to evaluate how well the trained models learn the object-centric grasp pose distribution. The SE(3) \acrshort*{emd} metric evaluates the empirical distance between two distributions of SE(3) poses. To compute this, we sample 100 grasps per object from a model and sample 100 random grasps from the ground truth datasets. Subsequently, we compute the minimum distances between the two sets using linear sum assignment and report the mean distance over 500 iterations.

\subsection{6-DoF grasp generation}
\label{sec:full-pc-generation}
Here, we evaluate the ability of GraspLDM models to learn the complex distribution of successful grasp poses on full object point clouds. For each object, we take 3 randomly sampled and augmented point clouds and generate 100 grasps on each. We aggregate the simulation results from all 300 grasp attempts into a \textit{mean grasp success rate} per object reported in Fig.~\ref{fig:full_pc_success_rate}. To assess the benefits of adding the diffusion model, the success rate results of Grasp\acrshort*{ldm} models are compared with the corresponding Grasp\acrshort*{vae} models, which is the base \acrshort*{vae} model with the same weights. For the baseline, we use the $SE(3)$ grasp diffusion model~\cite{urain2022se} (SE3-DiF-63C). To understand the scaling behavior, we present results from each network on 1C and 63C sets. Further, Fig.~\ref{fig:full_pc_emd} reports the \acrshort*{emd} metric that evaluates the ability to cover the distribution of grasps in the data.

For success rate, we outline two observations in Fig.~\ref{fig:full_pc_success_rate}. First, the \acrshort*{ldm} models consistently improve upon the base~\acrshort*{vae} models, demonstrating that the sample quality improves by using a diffusion model in~\acrshort*{vae}'s latent space. Second, the success rate performance of GraspLDM models scales better to a larger object set (63C). Comparing the median of the success rate on the test set in Fig.~\ref{fig:full_pc_success_rate}, GraspLDM-1C improves the median success rate to 88\% compared to 84\% of GraspVAE-1C. On the larger 63C set, the GraspLDM-63C model improves the median success rate to 78\% compared to 66\% from GraspVAE-63C. GraspLDM models also improve the interquartile range in each case. For the comparison against the baseline, SE3-DiF-1C model reports a median success rate of 89\% which is comparable to 88\% the GraspLDM-1C model. On the larger 63C set, we observe that the GraspLDM-63C model registers 78\% median success rate and distinctly higher interquartile range compared to SE3-DiF-63C. Therefore, GraspLDM models scale more favorably to larger object sets compared to the state-of-the-art diffusion models for this task. During the tests, we also observed that the models available from~\cite{urain2022se} do not hold out a test split and are trained on all objects. Therefore, the comparison to SE3-DiF models~\cite{urain2022se} is conservative, which emphasizes the performance of GraspLDM on withheld objects. 

\begin{figure}[t]
    \centering
        \includegraphics[width=\linewidth]{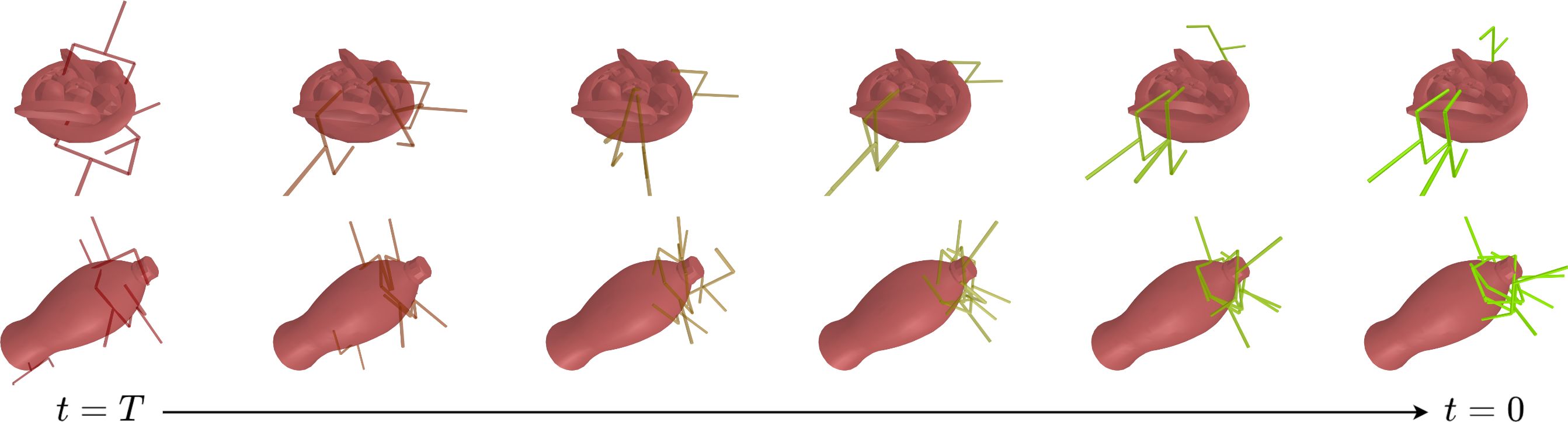}
         \caption{Visualization of latent space denoising in \mbox{GraspLDM}. Reverse diffusion de-noises these latents from $t=T$ to $t=0$, gradually moving the bad grasps towards regions of good grasps.}
         \label{fig:latent_dif_visualization}
\end{figure}%
~

Overall, the results validate our hypothesis that a diffusion model can bridge the prior gap in the latent space to provide higher-quality grasp pose samples. Fig.~\ref{fig:latent_dif_visualization} visualizes this effect where the latent denoising moves the latents corresponding to bad grasps towards good grasp regions during the reverse diffusion process. In terms of grasp failures, we observe a major portion of them in large objects whose graspable regions are far from the center of mass as shown in Fig.~\ref{fig:sim_failure}. When lifted in simulation, the reaction torque snaps the object out of the gripper fingers and therefore is reported as a failure.

Fig.~\ref{fig:full_pc_emd} shows that
our models also register low \acrshort*{emd} for most test objects, demonstrating good coverage of grasp modes in the ground truth data. The \acrshort*{emd} performance of GraspVAE and GraspLDM models are comparable or marginally better than the SE3-DiF models. However, the results show that while GraspLDM models effectively learn the grasp distribution, they have a slightly worse \acrshort*{emd} compared to the GraspVAE models. We attribute this to two factors. First, reverse diffusion moves poor grasp poses (e.g. colliding or free-space grasps) towards a smaller number of high-density regions that tend to provide a higher success rate. Second, we follow the computation of SE(3) EMD from~\cite{urain2022se} by using cosine distance for rotation and metric Euclidean distance for translation. As a result, the EMD metric is more sensitive to the similarity in rotation despite translation having a potentially larger effect on the success of grasp execution.

\begin{figure}[t]
    \centering
    \includegraphics[width=\linewidth]{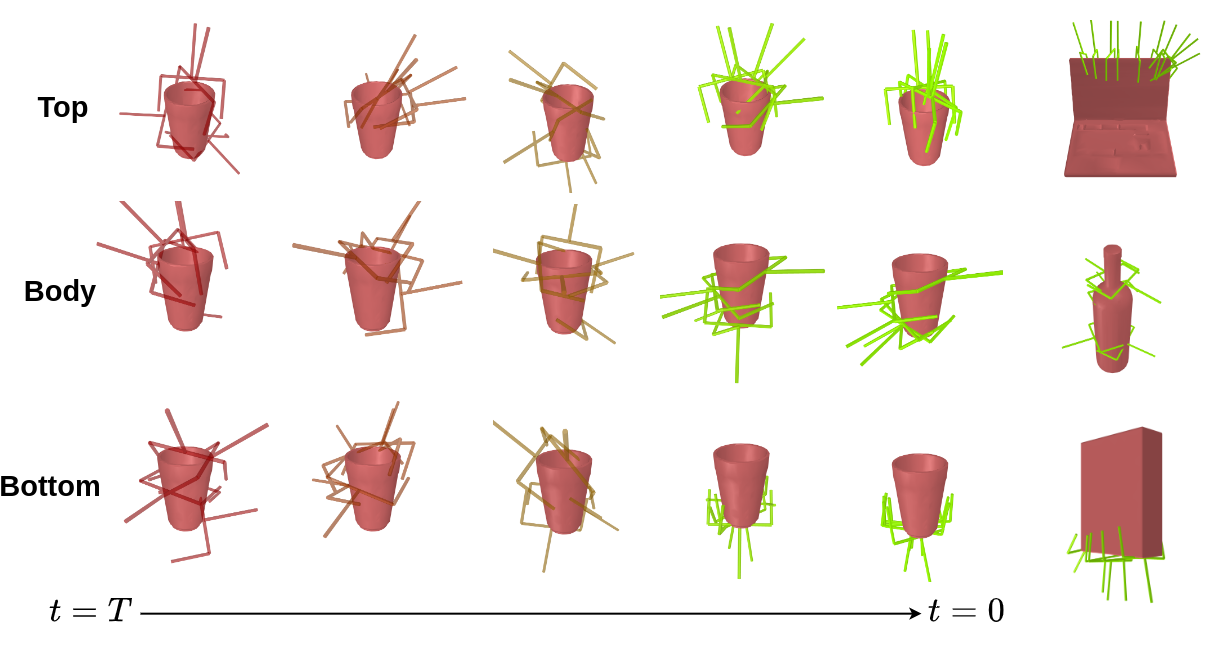}
        \caption{Task-conditional denoising in the latent space for a region-semantic class label (top, body, or bottom)}
    \label{fig:cond_gen}
\end{figure}

\subsection{Conditional generation}
\label{sec:results-condgen}
For many manipulation tasks, the desired grasps are subject to a task context. Here, we demonstrate the flexibility of our architecture to provide this task context as conditional guidance in the latent space representing unconditional grasps. We do this by training a task-conditional diffusion model post hoc. For the proof-of-concept, we consider simple region-semantic labels "top", "body" and "bottom" as our conditioning signals. We use full point clouds of objects as inputs and pre-train a \acrshort*{vae} without any task labels. We then include task labels only during the training of the diffusion model in the second stage. We associate a ground-truth class label of "top", "body" or "bottom" to each grasp based on whether the origin of the gripper's root is above, along, or below the unrotated object point cloud. We supply these labels to the diffusion model as an additional conditioning signal. We use a reduced subset of test objects for which these labels are meaningful (e.g. bottles) and remove those for which these labels are not relevant (e.g. plates). The final test set contains 200 unseen objects. We take pre-trained encoders and decoders from the GraspVAE-63C model and train the task-conditioned models (Task-GraspLDM) post-hoc, in less than 2 hours on a single NVIDIA V100 GPU.


\begin{table}[t]
    \centering
    \caption{Reverse diffusion sampling speed-up and performance of GraspLDM-63C. Standard DDPM (1000 steps) is compared with a fast sampler- DDIM (100 steps) for the number of grasps generated ($N_G$) without re-training.}
    \begin{tabular}{>{\centering\arraybackslash}p{10mm}|>{\centering\arraybackslash}p{15mm} >{\centering\arraybackslash}p{17mm}| >{\centering\arraybackslash}p{15mm}}
        \hline 
        Sampling & $N_G=100 $ & $N_G=1000 $ &  Median \\ 
         & (s) & (s) & success rate \\
         \hline 
        \acrshort{ddpm} & $7.39 \pm 0.06$  & $11.80 \pm 0.29$ & $\mathbf{0.792}$\\
        \acrshort{ddim} & $\mathbf{0.75 \pm 0.02}$ & $\mathbf{1.14 \pm 0.01}$ & $0.756$\\
        \hline
        \acrshort{vae} & $\mathbf{0.02 \pm 0.01}$ & $\mathbf{0.03 \pm 0.01}$ & $0.660$\\
        \hline
    \end{tabular}
    \label{tab:ddim_compare}
\end{table}

At test time, we generate 50 grasps for each mode (top, bottom, and body) of an object. The input object point cloud is augmented with random rotations. To check whether the generated grasp complies with the input label, we reverse the rotation transformation and check the location of the gripper's root, as earlier. We report the precision between the labels of generated grasps and the labels supplied as conditioning. In this setting, the Task-GraspLDM model provides a mean precision of 0.703 averaged over all objects. We observe that the reverse diffusion process seamlessly moves a latent from a prior distribution to the desired task-conditional distribution, as visualized in Fig.~\ref{fig:cond_gen}. These tests show that GraspLDM allows effective injection of task conditioning post-hoc. However, the models are limited by the cases where there are no ground-truth grasps for a given label. For instance, there are no 'bottom' grasp annotations for a laptop. Consequently, the model generates grasps unconditionally all over the object as shown in Fig.~\ref{fig:laptop_bottom_failure} disregarding the label.

~
\begin{figure}[t]
    \centering
        \centering
        \includegraphics[width=\linewidth]{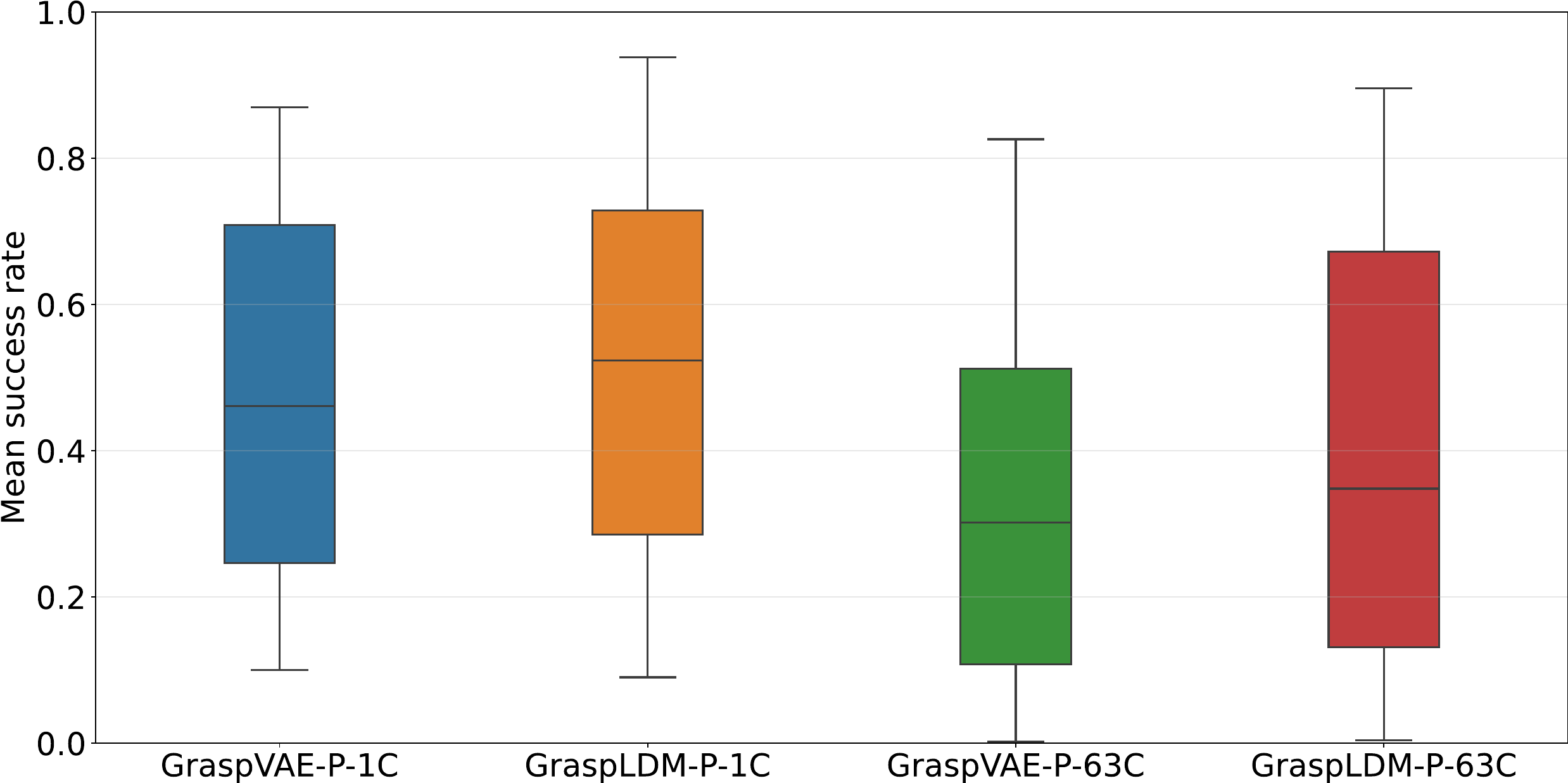} 
        \caption{Grasp generation performance on partial point clouds of 1C and 63C object sets.}
        \label{fig:partial_pc_success_rate}
        \vspace{-1\baselineskip}
\end{figure}

\subsection{Reverse diffusion sampling}
\label{sec:results-reverse-diff-sampler}
A notable disadvantage of using \acrshort*{ddpm}-based grasp generation is the time to execute large number reverse sampling steps sequentially. In \acrshort*{ddpm}, a large $T$ is required to ensure that the Gaussian conditional distributions in Eq.~\ref{eq:reverse_diff_kernel} are a good approximation~\cite{sohl2015deep}. A \acrlong{ddim}~\cite{song2020denoising} assumes a non-Markovian forward process to speed up the sample generation and requires lower sampling steps. Further, it can be used as a drop-in sampler for a score network trained on the \acrshort*{ddpm} objective in Eq.~\ref{eq:loss_diffusion}, without any re-training. To assess the performance and sampling speed trade-off, we take the GraspLDM-63C model and compare the 100-step \acrshort*{ddim} sampler against the naive 1000-step \acrshort*{ddpm} sampler. Table~\ref{tab:ddim_compare} compares the execution time of the reverse diffusion loop in \acrshort*{ddim} with the baseline 1000-step \acrshort*{ddpm}. Using \acrshort*{ddim} sampling as a drop-in replacement, the sampling time drops to 0.75s for 100 grasps and 1.1s for 1000 grasps. This is 10x faster than \acrshort*{ddpm} with a small loss in success rate (3.6\%). The experiment was conducted on a system with NVIDIA RTX 3080Ti GPU and Intel i7-12800H CPU. Given the rapid progress in creating fast reverse diffusion samplers, our pipeline provides the flexibility to trade off speed against performance for such diffusion samplers. In contrast, SE3-Grasp-DiF~\cite{urain2022se} models cannot leverage new samplers without re-training.    


     

\subsection{Single-view point clouds}
\label{sec:results-partial-pc}
For many real-world use cases, only single-view point clouds are available for grasp generation. Therefore, we also evaluate the GraspLDM framework's ability to learn the distribution of grasps on noisy partial point clouds. For evaluation, we use the simulation environment as before, with the addition of a depth camera to simulate real-world conditions where only partial object information is available. The cameras are spawned randomly between 30cm and 1m from the object. We feed the partial point cloud obtained from this single RGB-D view to the models. We generate 25 grasps per point cloud and repeat the generation for 20 random camera poses per object. Subsequently, we test the 500 grasps per object in the simulation environment. The success rate over the two object sets is reported in Fig.~\ref{fig:partial_pc_success_rate} for GraspVAE-P-1C, GraspLDM-P-1C, GraspVAE-P-63C, and GraspLDM-P-63C models, where 'P' implies that the model was trained on partial point clouds. In both cases, GraspLDM models improve the performance of the base \acrshort*{vae} model. We found that for GraspLDM-P-63C models, higher capacity is required to compress a meaningful latent representation of partial point clouds from the 63 categories. This is expected as the model learning from partial point cloud needs to additionally reason about the camera pose and shape completion aspects to make sense of the grasps and learn a structured latent space. We therefore increase the grasp latent ($\mathbf{z}_h$) size to 16 for GraspLDM-P-63C.   

\begin{figure}[t]
    \centering
     \begin{subfigure}[t]{0.3\linewidth}
        \centering
        \includegraphics[width=\linewidth]{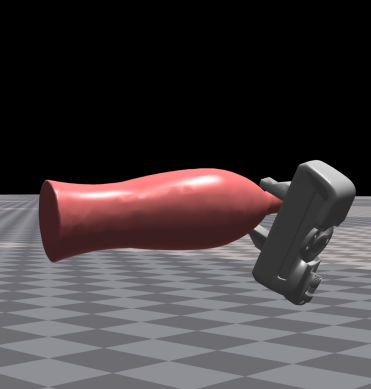}
        \caption{}
        \label{fig:sim_failure}
    \end{subfigure}~
    \begin{subfigure}[t]{0.3\linewidth}
        \centering
        \includegraphics[width=\textwidth]{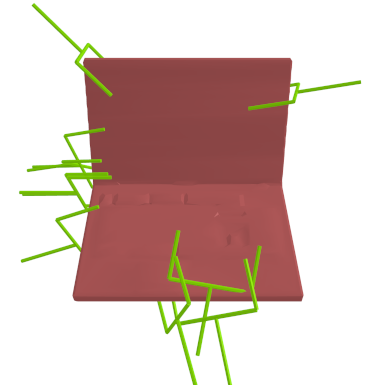}
        \caption{}
        \label{fig:laptop_bottom_failure}
    \end{subfigure}%
    ~
     \begin{subfigure}[t]{0.3\linewidth}
        \centering
        \includegraphics[width=\linewidth]{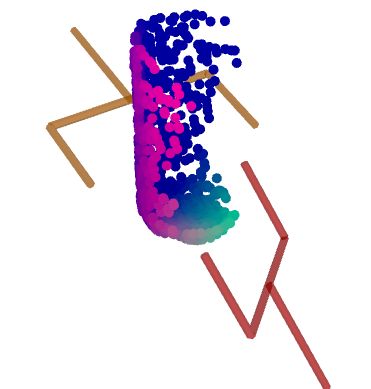}
        \caption{}
        \label{fig:partial_pc_failure}
    \end{subfigure}
\caption{ GraspLDM Limitations: (a) Failure on large objects from adverse torque. (b) Failure of region-semantic conditional generation for "bottom" grasps. (c) GraspLDM-P-63C generates colliding~(orange) and free-space grasps~(red).}
\end{figure}

Note that we do not use preferential camera viewpoints, grasp filtering, or a support surface in the background. As a result, this evaluation provides a conservative metric for 6-DoF grasp-generation performance. This can be considered as the approximate lower bound of the grasp generation performance for real-world use cases. A large number of failures result from grasps that collide with parts of the objects not visible in the partial point clouds. Depending on the viewpoint, this occlusion confuses the model to consider incomplete surfaces as edges along which the object could be grasped, as shown in Fig.~\ref{fig:partial_pc_failure}. For larger objects where grasps are concentrated in a small region of the object, another issue affecting the performance is the visibility of these regions in the point cloud.


\subsection{Real robot tests}
\label{sec:results-real-world}
To complement the large-scale evaluations conducted in simulation, we conduct thorough real-world tests to evaluate the performance of object-centric GraspLDM models. We compare our performance against two established baselines widely used in the community. First, as a direct comparison with a generative model architecture, we use 6-DoF-Graspnet~\cite{mousavian20196}. Second, we compare the effectiveness of our models against the state-of-the-art Contact-Graspnet~\cite{sundermeyer2021contact}, which uses a feed-forward PointNet style architecture. We do not use SE3-DiF diffusion models~\cite{urain2022se} for real-world tests as the training and testing in the original work are conducted with full object point clouds. A fair and reliable comparison with these models was not possible as we observed significant performance degradation when training SE3-DiF models on single-view point clouds from 63 categories with the default hyperparameters. 

\begin{figure}[t]
    \centering
    \includegraphics[width=\linewidth]{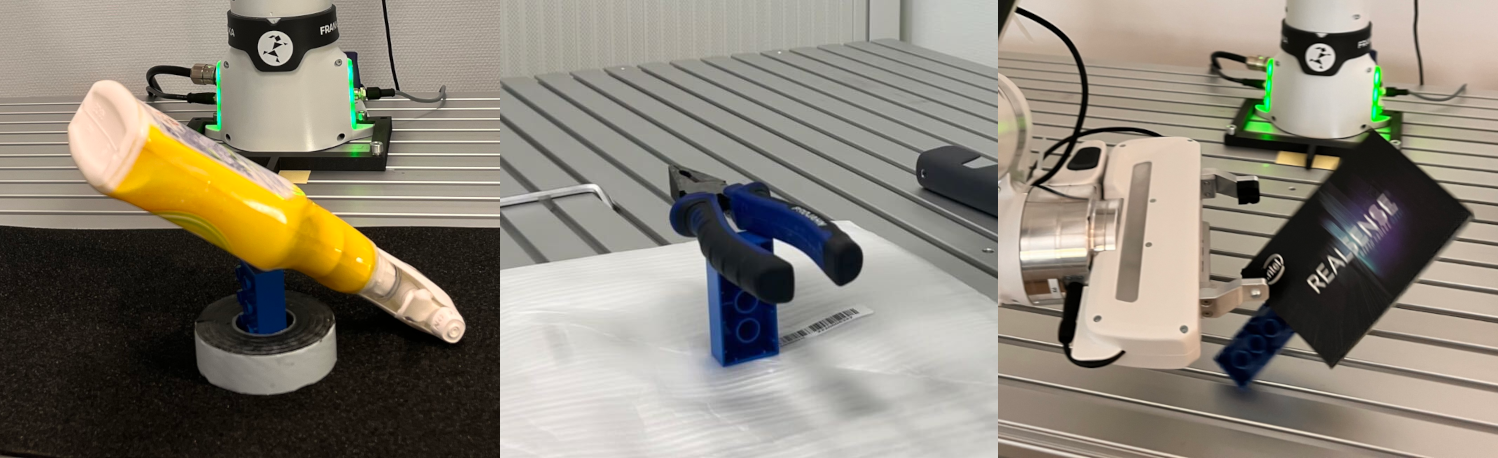}
    \caption{Examples of object placement on support during real-world testing. Such object placements allow grasps from more approach directions to be feasible while also demanding higher grasp pose accuracy for success.}  \label{fig:object_placement}
\end{figure}
\begin{figure*}[ht]
    \centering
        \centering
        \includegraphics[width=\linewidth]{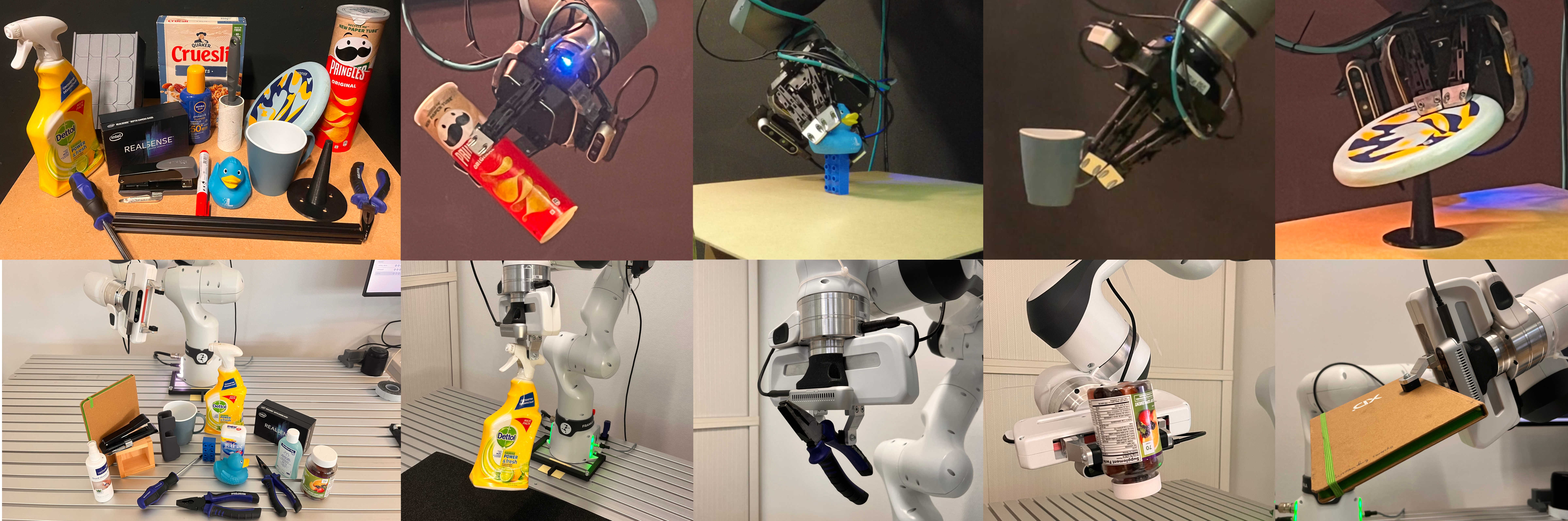} 
        \caption{Real-world validation of GraspLDM models with objects of diverse visual, geometric, and surface properties across two different robotic setups.}
        \label{fig:test_objs}
\end{figure*}

We assess the sim-to-real transfer on two hardware setups. The first setup uses a wall-mounted 6-DoF UR-10e arm, a Robotiq-3F gripper, and an Intel Realsense D435 RGB-D camera. The second setup uses a table-mounted 7-DoF Franka Research 3 arm with Franka hand. In the former case, the 3-finger gripper is operated in a two-finger "pinch" mode with the gripper sweep length restricted to 85mm to emulate the gripper used for generating the training data. In both cases, the camera is mounted at the tool flange of the arm in an eye-in-hand configuration. Due to limitations of access to hardware, we conducted the comparison with 6-DoF Graspnet on the first setup, while the comparison with Contact-Graspnet was conducted later on the second setup. In each case, we use a test set of 16 randomly selected objects of diverse physical and geometric properties as shown in Fig.~\ref{fig:test_objs}. Each grasping trial is conducted by placing a single object in a random pose on a table. To ensure that the trials allow execution of grasps from more approach directions than just top-down, we lift the objects on a small support as shown in Fig. ~\ref{fig:object_placement}. This also makes grasping success more difficult compared to table-top placement where in-plane sliding and normal contact force supports imprecise grasp attempts. At the start of the trial, the robot observes the object from an arbitrarily fixed pose such that the camera boresight is between 30$^\circ$ and 60$^\circ$ to the table plane and the object is between 0.4m and 0.8m from the camera. We conduct five random pose trials per object to evaluate the success rate across a total of 80 grasp attempts without retries.

~
\begin{table*}[t]
    \centering
    \caption{Real-world 6-DoF grasping success rate comparison on 16 evaluation objects in 5 random poses}
    \begin{tabularx}{0.7\linewidth}{p{4.5cm}|p{5.5cm}|c}
    \hline 
    \textbf{Setup} & \textbf{Method} & \textbf{Success} \\ 
    & & \textbf{Rate}   \\
    \hline 
    \multirow{3}{*}{Setup 1: UR10e + Robotiq-3F}
    & \textbf{GraspLDM+GraspClassifier} \textit{(ours)} & $\mathbf{80}$\% \\
    & GraspVAE+GraspClassifier \textit{(ours)} & $76.25$\% \\
    & 6-DoF-Graspnet+Classifier~\cite{mousavian20196} & $37.5$\% \\
    \hline 
    \multirow{3}{*}{Setup 2: Franka FR3 + Franka hand}
    & \textbf{GraspLDM+GraspClassifier} \textit{(ours)} & $\mathbf{78.75}$\% \\
    & GraspVAE+GraspClassifier \textit{(ours)} & $67.5$\% \\
    & Contact-Graspnet~\cite{sundermeyer2021contact} & $76.25$\% \\ 
    \hline
    \end{tabularx}
    \label{tab:real_test_success_rate}
\end{table*}
We use the GraspLDM-P-63C model introduced in Section~\ref{sec:results-partial-pc} without any further fine-tuning. Since the models provide object-centric grasps, we use the segment-anything model~\cite{kirillov2023segment} to segment the object in an RGB image. Subsequently, we mask the aligned depth image to provide a segmented object point cloud. This point cloud is provided to the model which generates 100 grasp candidates. Unlike simulation, real experiments require a grasp pose to be selected. For this purpose, we train a naive classifier (GraspClassifier) to provide a pseudo-probability of success for a given pair of a point cloud and a grasp pose. Subsequently, we rank and sort the grasps above a probability threshold ($p >0.5$). We attempt the grasps in the same order while checking for collision and kinematic failures. The first grasp with a valid plan is executed on the robot. For motion planning, we utilize an off-the-shelf RRTConnect planner and plan cartesian trajectories to the grasp pose. During a grasp attempt, the robot first goes to a pre-grasp pose, which is offset from the grasp pose along the z-axis of the gripper. Then the gripper approaches the object along this axis and closes the fingers. Subsequently, the object is lifted and moved to a set pose above a bucket. A grasp is successful if all these steps are executed while keeping the object attached to the gripper. Together these steps test the validity of the grasp pose and the stability of the grasp in all axes. 
For a fair evaluation of grasp synthesis models, we only use the generator and classifier from 6-DoF-Graspnet and do not include the iterative refinement stage. Similarly, we do not perform iterative refinement on top of Contact-Graspnet predictions. All other parts of the pipeline like segmentation remain the same in every case. We compare this with GraspLDM-P-63C and GraspVAE-P-63C in Table~\ref{tab:real_test_success_rate}. 

Table~\ref{tab:real_test_success_rate} demonstrates that GraspLDM provides a superior grasp success rate in the real world while being entirely trained with simulation data. Compared to the generative model baseline~\cite{mousavian20196}, GraspLDM demonstrates a significantly higher success rate. On the other hand, GraspLDM shows a comparable success rate with that of Contact-Graspnet~\cite{sundermeyer2021contact}, which is a feed-forward predictive model. Despite the success rate being marginally higher for GraspLDM, the difference concerns 2/80 grasp attempts. Due to several factors influencing real-world tests of 6-DoF grasps, we consider the two models to have comparable success rates. We also observe that the GraspLDM-P-63C model provides around 4.75\% and 11.25\% higher success rate over its base~\acrshort*{vae} model in the two test setups while using the same classifier. We believe that the relative variation in the performance of GraspVAE between the two test setups is mainly related to the non-trivial effects of viewpoint and noise on generation. We observe that for favorable viewpoints both GraspVAE and GraspLDM have similar output, while in some other cases, GraspLDM performs noticeably better than GraspVAE. This effect is more pronounced in the second setup than in the first. 

We notice that high amounts of failure for the 6-DoF-Graspnet models result from the grasps colliding with the object and a lack of grasp generation diversity. In the latter case, it does not produce enough grasps in kinematically feasible regions of the object in a given pose. We also observe a higher rate of planning failures for this baseline until a successful grasp is attempted, which is connected to grasp diversity. On the other hand, Contact-Graspnet outperforms and produces superior grasps for objects closer to the categories represented in the ACRONYM dataset. This is especially the case for objects with trivial edges/surfaces like boxes. While for the objects with curvature and those with shapes unlike ACRONYM categories, it underperforms compared to GraspLDM. Overall, we demonstrate that GraspLDM models outperform existing generative models and provide comparable performance against larger feed-forward models.


\section{Limitations and Future Work}

GraspLDM demonstrates that latent diffusion provides a powerful mechanism to improve generative performance, flexibility to downstream tasks, and sample quality for real-world grasping. However, this work is not without limitations. The primary limitation comes from not being able to bootstrap a notion of grasp quality. This currently necessitates a separate classifier in practical applications. In the real-world tests, we train a naive GraspClassifier, which is over-confident and prefers deeper grasp penetration around the object. An external classifier like this also hinders the framework's flexibility to enable plug-and-play task-specific diffusion models which need to reason about the grasps that are stable and suitable for the task at the same time. Therefore, some notion of grasp quality should be systematically bootstrapped inside the GraspLDM architecture in the future. 

Furthermore, our models are competitive despite being trained on only 63 out of 180 categories in the ACRONYM dataset due to compute constraints. We believe that the performance can be improved further by incorporating the entire dataset in the training. Finally, by training the latent diffusion model inside a frozen latent space of an unconditional GraspVAE, our framework paves the way for a flexible plug-and-play approach to task-specific grasping that inherits from a general grasping model. However, the analysis needs to be extended to more complex task conditioning inputs like points, poses, and embeddings. 

\section{Conclusions}
Learning generalizable representations is fundamental to robotic grasping in the real world. In this work, we propose \acrshort*{gldm}—a novel generative framework to learn object-centric representations for 6-DoF grasp synthesis. Our architecture allows denoising diffusion models to be used as expressive priors in the latent space of \acrshort*{vae}s. This enables efficient learning of the complex distributions of 6-DoF grasp poses on object-centric point clouds. We conduct large-scale simulation tests to show that GraspLDM outperforms baseline methods and provides a 78\% median success rate on our test set of 400 objects from 63 categories. GraspLDM demonstrates better grasp generation performance while scaling favorably to large object sets compared to existing generative models. We also demonstrate GraspLDM's ability to efficiently train task-specific models and use fast reverse diffusion samplers for downstream applications. Further, we validate our pipeline by training GraspLDM models on single-view point clouds generated in simulation and testing them on two different real-world setups. GraspLDM demonstrates effective sim-to-real transfer, achieving 78\% and 80\% success rates across 80 grasp attempts on 16 diverse objects in two different real-world robotic setups. GraspLDM outperforms contemporary generative models and matches the performance of state-of-the-art non-generative models while offering greater flexibility.


\bibliographystyle{IEEEtran.bst}
\bibliography{references}
\vspace{5\baselineskip}
\begin{IEEEbiography}[{\includegraphics[width=1in,height=1.25in,clip,keepaspectratio]{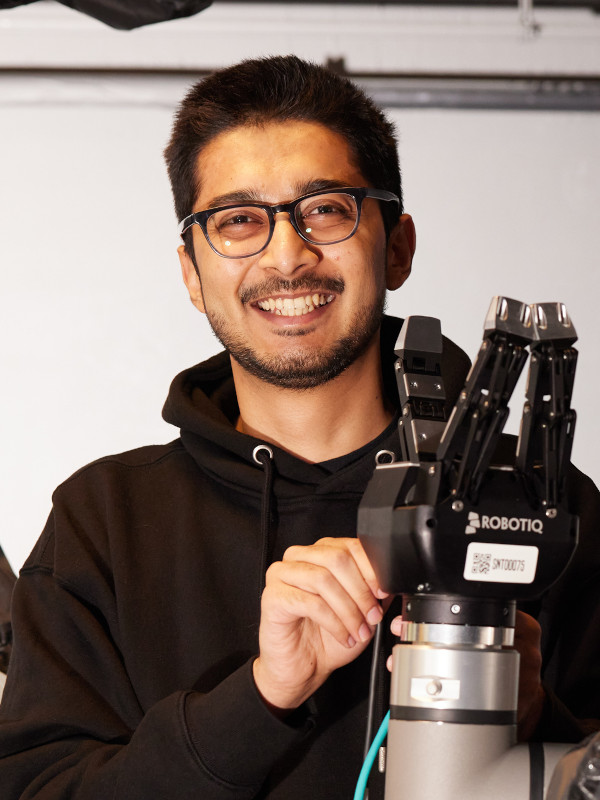}}]{KULDEEP R BARAD} (Member,
IEEE) received his B.Tech (distinction) and M.Sc (cum laude) in Aerospace Engineering from SRM University and Delft University of Technology respectively. Currently, Kuldeep is a PhD candidate in the Space Robotics Research Group at SnT-University of Luxembourg. He was awarded an Industrial Fellowship grant from the Luxembourg National Research Fund to support his doctoral research. As a part of this fellowship, he conducts research jointly with Redwire Space Europe on advancing perception for robotic manipulation systems in space.     
\end{IEEEbiography}

\begin{IEEEbiography}[{\includegraphics[width=1in,height=1.25in,clip,keepaspectratio]{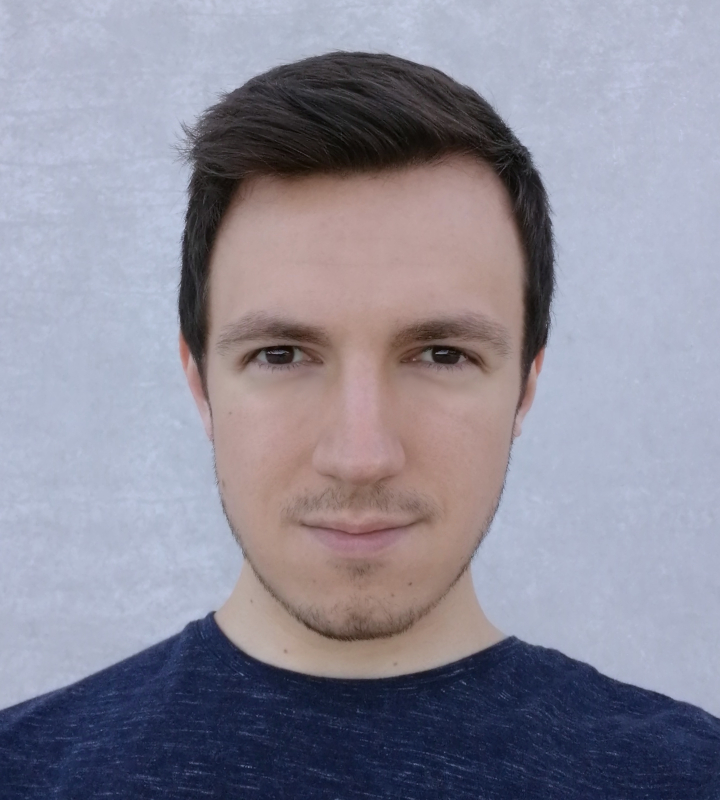}}]{ANDREJ ORSULA} received his BSc and MSc degrees in Robotics from Aalborg University in 2019 and 2021, respectively. He is currently a PhD candidate in the Space Robotics Research Group (SpaceR) at the University of Luxembourg, where his research focuses on advancing the application of learning-based techniques for robotic assembly and servicing in space.
\end{IEEEbiography}

\begin{IEEEbiography}
[{\includegraphics[width=1in,height=1.25in,clip,keepaspectratio]{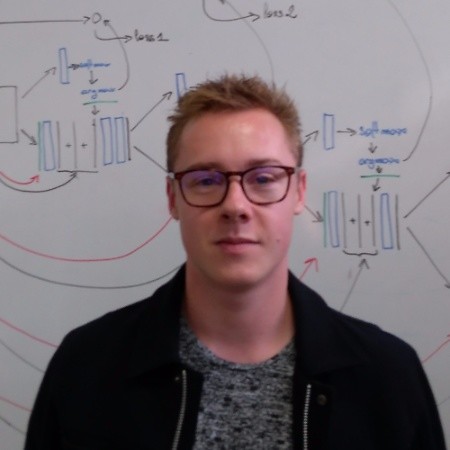}}] {ANTOINE RICHARD} received his Ph.D. from the Georgia Institute of Technology (a.k.a. GeorgiaTech) in 2022. During his thesis, he studied the modeling and control of dynamic systems through the use of neural networks and reinforcement learning. Antoine Richard is now a Research Associate in the Space Robotics (SpaceR) Research Group at the University of Luxembourg, where he works on Environment Generation and Reinforcement Learning for Space Robotics.

\end{IEEEbiography}

\begin{IEEEbiography}[{\includegraphics[width=1in,height=1.25in,clip,keepaspectratio]{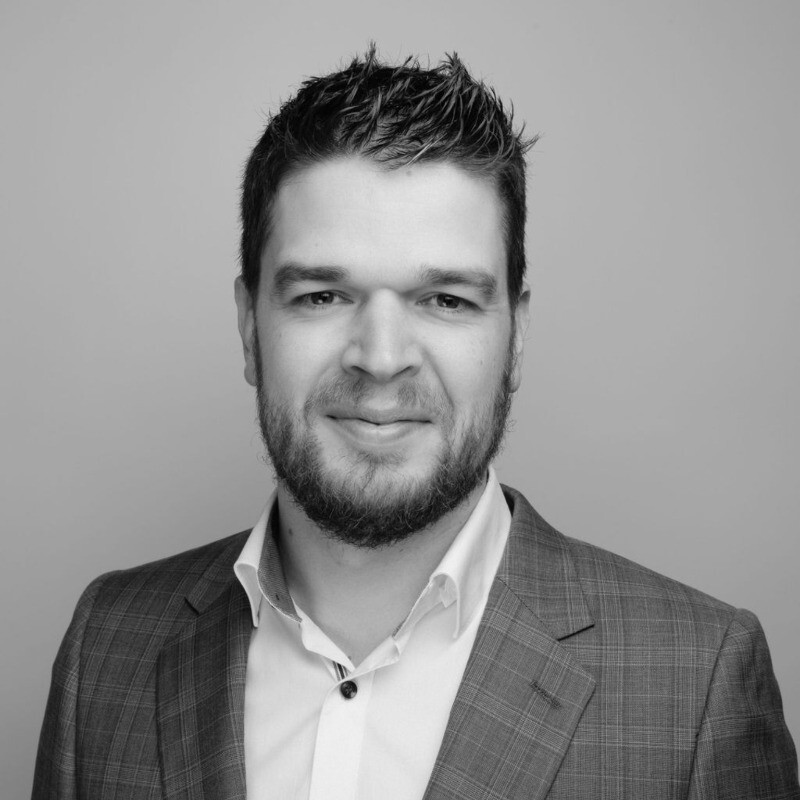}}]{JAN DENTLER} is the Research and Development manager at Redwire Space Luxembourg, where he is responsible for the technology roadmap. He further leads the development of software, avionics, and high-level control for space robotics systems.
Before Redwire Space Europe, he completed his Ph.D. (Excellent) in Control Engineering at the Automation and Robotics research group of the University of Luxembourg. In his doctoral dissertation- "Real-time model predictive control of cooperative aerial manipulation", he worked on manipulation for volatile robotic platforms. He has been the first author of multiple international publications on robotic control. 
Prior to this, he received an M.Sc. in automation and control from Ulm University, a Bachelor of Engineering with honors from the University of Applied Science Ulm, and a technician degree with honors in industrial electronics.
His expertise is robotics, software architecture, nonlinear control, real-time optimization, system modeling, identification, and applied mathematics.
\end{IEEEbiography}

\begin{IEEEbiography}[{\includegraphics[width=1in,height=1.25in,clip,keepaspectratio]{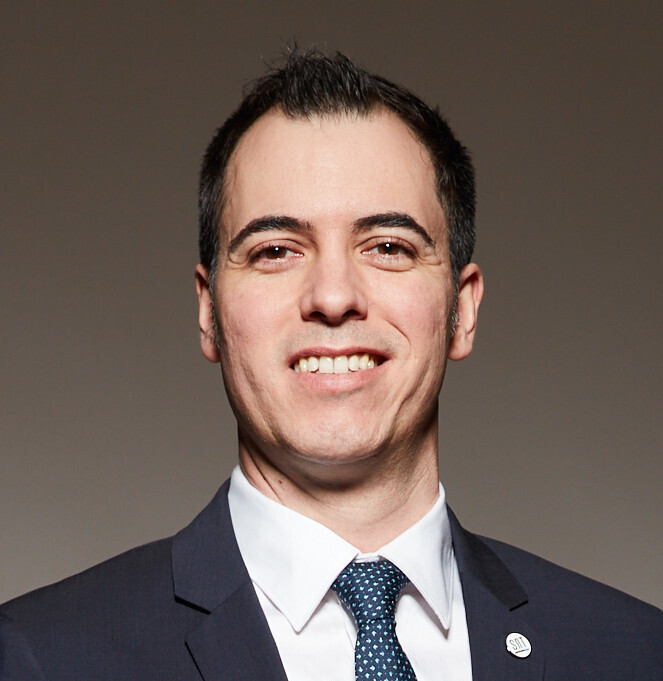}}]{MIGUEL A. OLIVARES-MENDEZ} (Member,
IEEE) received the Engineering degree in computer science from the University of Malaga, in 2006, and the M.Sc. and Ph.D. degrees in robotics and automation from the Technical University of Madrid, in 2009 and 2013, respectively.
During his Ph.D., he was a visiting researcher
with EPFL, Switzerland, and ARCAA-QUT,
Australia. In May 2013, he joined the Interdisciplinary Centre for Security Reliability and Trust
(SnT), University of Luxembourg as a Research Associate
with the Automation and Robotics Research Group (ARG). In December 2016,
he was appointed as a Research Scientist, responsible for the research
activities on mobile robotics at ARG-SnT. He is currently a tenured assistant professor of
space robotics and a senior research scientist at SnT and the University of Luxembourg.
He leads the Space Robotics Research Group (SpaceR) where he is the main supervisor for 11 Ph.D. students and 7 post-doctoral researchers. He has published more than 120 peer-reviewed publications. His research interests include aerial, planetary and orbital robotics for autonomous navigation, situational awareness, perception,
machine learning, multi-robot interaction in autonomous exploration,
inspection, and operations.
Dr. Olivares-Mendez is an Associate Editor for IROS, ICRA, and ICUAS
conferences; and the Journal of Intelligent and Robotics Systems, Frontiers
on Space and Field Robotics, and The International Journal of Robotics
Research (IJRR).
\end{IEEEbiography}

\begin{IEEEbiography}[{\includegraphics[width=1in,height=1.25in,clip,keepaspectratio]{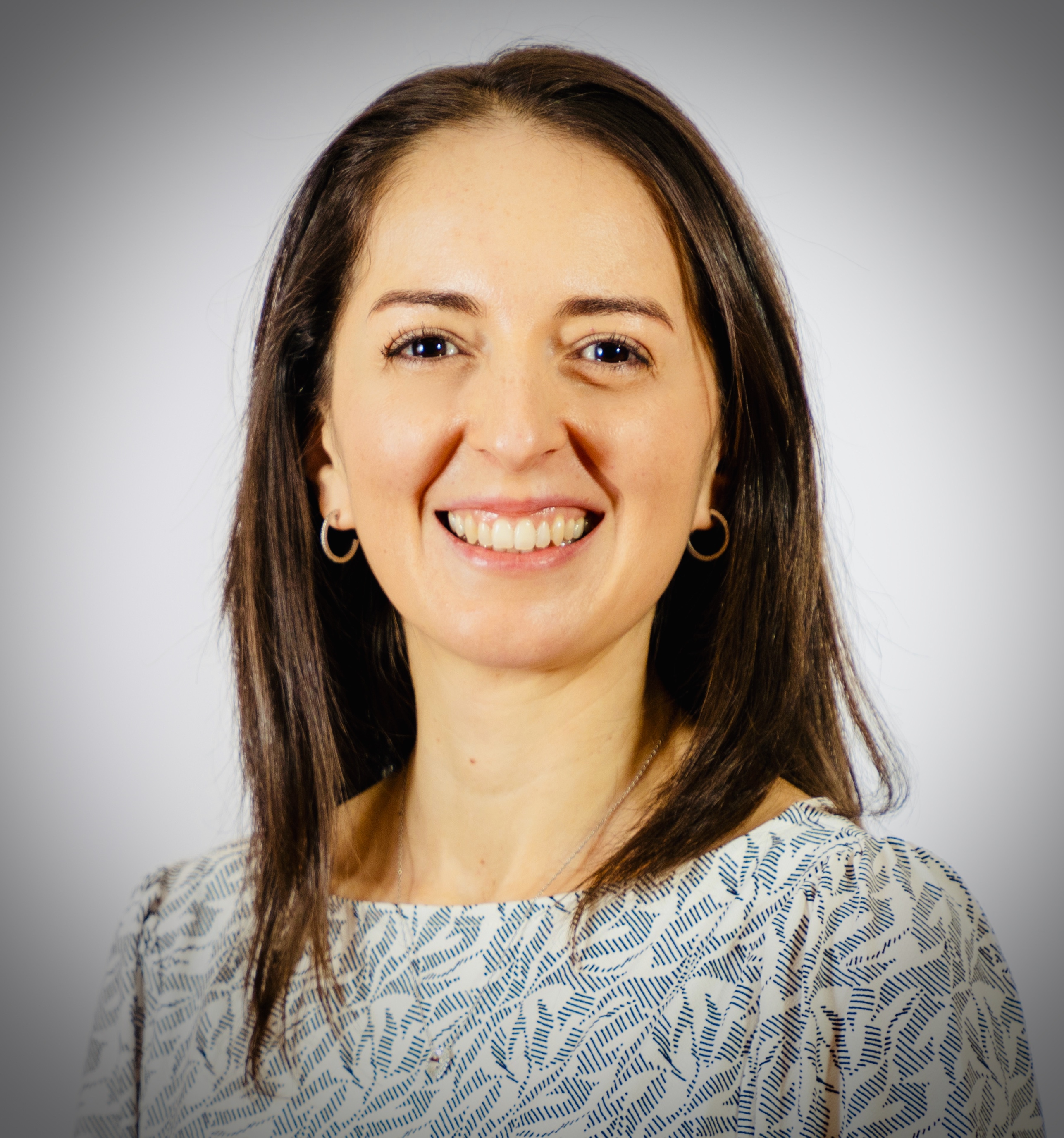}}]{CAROL MARTINEZ} received
her M.Sc. and Ph.D. degrees in robotics and
automation from Universidad Politécnica de
Madrid (UPM), in 2009 and 2013, respectively,
with a focus on computer vision for unmanned
aerial vehicles (visual tracking, pose estimation,
and control), for which she received the outstanding Ph.D. thesis award by UPM. As a Ph.D.
candidate, she was a Visiting Researcher with
the Queensland University of Technology and the
University of Bristol, U.K., where she developed algorithms for tracking and
pose estimation using cameras on-board aerial vehicles. She held positions
as a Postdoctoral Researcher with UPM, and an Assistant Professor with
PUJ, Bogotá, Colombia, from 2015 to 2020. She has led and conducted
interdisciplinary research in computer vision, machine learning, and deep
learning for process automation (industry and health) and robotics (aerial,
industrial, and space). Since 2020, she has been a Research Scientist with the Space Robotics Research Group at SnT-University of Luxembourg. She leads projects in orbital robotics, robotic manipulation,
and on-ground emulation. 
Her research interests include perception approaches for autonomous operation of robots in space and multi-purpose manipulation tasks for
planetary and orbital robotics applications.
\end{IEEEbiography}

\newpage


\EOD

\end{document}

%% file: acronym.tex
\newacronym{vae}{VAE}{Variational Autoencoder}
\newacronym{ldm}{LDM}{Latent Diffusion Model}
\newacronym{ddm}{DDM}{Denoising Diffusion Model}
\newacronym{cvae}{CVAE}{Conditional Variational Autoencoder}
\newacronym{gldm}{GraspLDM}{Grasp Latent Diffusion Models}
\newacronym{emd}{EMD}{Earth Mover's Distance}
\newacronym{mrp}{MRP}{Modified Rodrigues Parameters}
\newacronym{edm}{EDM}{Elucidated Diffusion Model}
\newacronym{sde}{SDE}{Stochastic Differential Equations}
\newacronym{ode}{ODE}{Ordinary Differential Equations}
\newacronym{sgm}{SGM}{Score-based Generative Model}
\newacronym{ddpm}{DDPM}{Denoising Diffusion Probabilistic Models}
\newacronym{elbo}{ELBO}{Evidence Lower Bound}
\newacronym{ddim}{DDIM}{Denoising Diffusion Implicit Model}
\newacronym{dof}{DoF}{Degrees of Freedom}

%% file: access.bbl
\begin{thebibliography}{10}
\providecommand{\url}[1]{#1}
\csname url@samestyle\endcsname
\providecommand{\newblock}{\relax}
\providecommand{\bibinfo}[2]{#2}
\providecommand{\BIBentrySTDinterwordspacing}{\spaceskip=0pt\relax}
\providecommand{\BIBentryALTinterwordstretchfactor}{4}
\providecommand{\BIBentryALTinterwordspacing}{\spaceskip=\fontdimen2\font plus
\BIBentryALTinterwordstretchfactor\fontdimen3\font minus \fontdimen4\font\relax}
\providecommand{\BIBforeignlanguage}[2]{{%
\expandafter\ifx\csname l@#1\endcsname\relax
\typeout{** WARNING: IEEEtran.bst: No hyphenation pattern has been}%
\typeout{** loaded for the language `#1'. Using the pattern for}%
\typeout{** the default language instead.}%
\else
\language=\csname l@#1\endcsname
\fi
#2}}
\providecommand{\BIBdecl}{\relax}
\BIBdecl

\bibitem{bohg2013data}
J.~Bohg, A.~Morales, T.~Asfour, and D.~Kragic, ``Data-driven grasp synthesis—a survey,'' \emph{IEEE Transactions on robotics}, vol.~30, no.~2, pp. 289--309, 2013.

\bibitem{bicchi2000robotic}
A.~Bicchi and V.~Kumar, ``Robotic grasping and contact: A review,'' in \emph{Proceedings 2000 ICRA. Millennium conference. IEEE international conference on robotics and automation. Symposia proceedings (Cat. No. 00CH37065)}, vol.~1.\hskip 1em plus 0.5em minus 0.4em\relax IEEE, 2000, pp. 348--353.

\bibitem{balasubramanian2012physical}
R.~Balasubramanian, L.~Xu, P.~D. Brook, J.~R. Smith, and Y.~Matsuoka, ``Physical human interactive guidance: Identifying grasping principles from human-planned grasps,'' \emph{IEEE Transactions on Robotics}, vol.~28, no.~4, pp. 899--910, 2012.

\bibitem{detry2012generalizing}
R.~Detry, C.~H. Ek, M.~Madry, J.~Piater, and D.~Kragic, ``Generalizing grasps across partly similar objects,'' in \emph{2012 IEEE International Conference on Robotics and Automation}.\hskip 1em plus 0.5em minus 0.4em\relax IEEE, 2012, pp. 3791--3797.

\bibitem{el2008handling}
S.~El-Khoury and A.~Sahbani, ``Handling objects by their handles,'' in \emph{IEEE/RSJ International Conference on Intelligent Robots and Systems}, no. POST\_TALK, 2008.

\bibitem{jiang2011efficient}
Y.~Jiang, S.~Moseson, and A.~Saxena, ``Efficient grasping from rgbd images: Learning using a new rectangle representation,'' in \emph{2011 IEEE International conference on robotics and automation}.\hskip 1em plus 0.5em minus 0.4em\relax IEEE, 2011, pp. 3304--3311.

\bibitem{mahler2017dex}
J.~Mahler, J.~Liang, S.~Niyaz, M.~Laskey, R.~Doan, X.~Liu, J.~A. Ojea, and K.~Goldberg, ``Dex-net 2.0: Deep learning to plan robust grasps with synthetic point clouds and analytic grasp metrics,'' \emph{arXiv preprint arXiv:1703.09312}, 2017.

\bibitem{morrison2018closing}
D.~Morrison, P.~Corke, and J.~Leitner, ``Closing the loop for robotic grasping: A real-time, generative grasp synthesis approach,'' \emph{arXiv preprint arXiv:1804.05172}, 2018.

\bibitem{newbury2022deep}
R.~Newbury, M.~Gu, L.~Chumbley, A.~Mousavian, C.~Eppner, J.~Leitner, J.~Bohg, A.~Morales, T.~Asfour, D.~Kragic \emph{et~al.}, ``Deep learning approaches to grasp synthesis: A review,'' \emph{arXiv preprint arXiv:2207.02556}, 2022.

\bibitem{ibarz2021train}
J.~Ibarz, J.~Tan, C.~Finn, M.~Kalakrishnan, P.~Pastor, and S.~Levine, ``How to train your robot with deep reinforcement learning: lessons we have learned,'' \emph{The International Journal of Robotics Research}, vol.~40, no. 4-5, pp. 698--721, 2021.

\bibitem{eppner2021acronym}
C.~Eppner, A.~Mousavian, and D.~Fox, ``Acronym: A large-scale grasp dataset based on simulation,'' in \emph{2021 IEEE International Conference on Robotics and Automation (ICRA)}.\hskip 1em plus 0.5em minus 0.4em\relax IEEE, 2021, pp. 6222--6227.

\bibitem{fang2020graspnet}
H.-S. Fang, C.~Wang, M.~Gou, and C.~Lu, ``Graspnet-1billion: A large-scale benchmark for general object grasping,'' in \emph{Proceedings of the IEEE/CVF conference on computer vision and pattern recognition}, 2020, pp. 11\,444--11\,453.

\bibitem{mousavian20196}
A.~Mousavian, C.~Eppner, and D.~Fox, ``6-dof graspnet: Variational grasp generation for object manipulation,'' in \emph{Proceedings of the IEEE/CVF International Conference on Computer Vision}, 2019, pp. 2901--2910.

\bibitem{murali20206}
A.~Murali, A.~Mousavian, C.~Eppner, C.~Paxton, and D.~Fox, ``6-dof grasping for target-driven object manipulation in clutter,'' in \emph{2020 IEEE International Conference on Robotics and Automation (ICRA)}.\hskip 1em plus 0.5em minus 0.4em\relax IEEE, 2020, pp. 6232--6238.

\bibitem{kingma2013auto}
D.~P. Kingma and M.~Welling, ``Auto-encoding variational bayes,'' \emph{arXiv preprint arXiv:1312.6114}, 2013.

\bibitem{ho2020denoising}
J.~Ho, A.~Jain, and P.~Abbeel, ``Denoising diffusion probabilistic models,'' \emph{Advances in Neural Information Processing Systems}, vol.~33, pp. 6840--6851, 2020.

\bibitem{song2019generative}
Y.~Song and S.~Ermon, ``Generative modeling by estimating gradients of the data distribution,'' \emph{Advances in neural information processing systems}, vol.~32, 2019.

\bibitem{sohl2015deep}
J.~Sohl-Dickstein, E.~Weiss, N.~Maheswaranathan, and S.~Ganguli, ``Deep unsupervised learning using nonequilibrium thermodynamics,'' in \emph{International Conference on Machine Learning}.\hskip 1em plus 0.5em minus 0.4em\relax PMLR, 2015, pp. 2256--2265.

\bibitem{song2020score}
Y.~Song, J.~Sohl-Dickstein, D.~P. Kingma, A.~Kumar, S.~Ermon, and B.~Poole, ``Score-based generative modeling through stochastic differential equations,'' \emph{arXiv preprint arXiv:2011.13456}, 2020.

\bibitem{zeng2022lion}
X.~Zeng, A.~Vahdat, F.~Williams, Z.~Gojcic, O.~Litany, S.~Fidler, and K.~Kreis, ``Lion: Latent point diffusion models for 3d shape generation,'' \emph{arXiv preprint arXiv:2210.06978}, 2022.

\bibitem{zhao2017infovae}
S.~Zhao, J.~Song, and S.~Ermon, ``Infovae: Information maximizing variational autoencoders,'' \emph{arXiv preprint arXiv:1706.02262}, 2017.

\bibitem{rombach2022high}
R.~Rombach, A.~Blattmann, D.~Lorenz, P.~Esser, and B.~Ommer, ``High-resolution image synthesis with latent diffusion models,'' in \emph{Proceedings of the IEEE/CVF Conference on Computer Vision and Pattern Recognition}, 2022, pp. 10\,684--10\,695.

\bibitem{urain2022se}
J.~Urain, N.~Funk, G.~Chalvatzaki, and J.~Peters, ``Se (3)-diffusionfields: Learning cost functions for joint grasp and motion optimization through diffusion,'' \emph{arXiv preprint arXiv:2209.03855}, 2022.

\bibitem{ekvall2007learning}
S.~Ekvall and D.~Kragic, ``Learning and evaluation of the approach vector for automatic grasp generation and planning,'' in \emph{Proceedings 2007 IEEE International Conference on Robotics and Automation}.\hskip 1em plus 0.5em minus 0.4em\relax IEEE, 2007, pp. 4715--4720.

\bibitem{ciocarlie2014towards}
M.~Ciocarlie, K.~Hsiao, E.~G. Jones, S.~Chitta, R.~B. Rusu, and I.~A. {\c{S}}ucan, ``Towards reliable grasping and manipulation in household environments,'' in \emph{Experimental Robotics: The 12th International Symposium on Experimental Robotics}.\hskip 1em plus 0.5em minus 0.4em\relax Springer, 2014, pp. 241--252.

\bibitem{sundermeyer2021contact}
M.~Sundermeyer, A.~Mousavian, R.~Triebel, and D.~Fox, ``Contact-graspnet: Efficient 6-dof grasp generation in cluttered scenes,'' in \emph{2021 IEEE International Conference on Robotics and Automation (ICRA)}.\hskip 1em plus 0.5em minus 0.4em\relax IEEE, 2021, pp. 13\,438--13\,444.

\bibitem{jiang2021synergies}
Z.~Jiang, Y.~Zhu, M.~Svetlik, K.~Fang, and Y.~Zhu, ``Synergies between affordance and geometry: 6-dof grasp detection via implicit representations,'' \emph{arXiv preprint arXiv:2104.01542}, 2021.

\bibitem{li2022gendexgrasp}
P.~Li, T.~Liu, Y.~Li, Y.~Geng, Y.~Zhu, Y.~Yang, and S.~Huang, ``Gendexgrasp: Generalizable dexterous grasping,'' \emph{arXiv preprint arXiv:2210.00722}, 2022.

\bibitem{liu2019point}
Z.~Liu, H.~Tang, Y.~Lin, and S.~Han, ``Point-voxel cnn for efficient 3d deep learning,'' \emph{Advances in Neural Information Processing Systems}, vol.~32, 2019.

\bibitem{perez2018film}
E.~Perez, F.~Strub, H.~De~Vries, V.~Dumoulin, and A.~Courville, ``Film: Visual reasoning with a general conditioning layer,'' in \emph{Proceedings of the AAAI Conference on Artificial Intelligence}, vol.~32, no.~1, 2018.

\bibitem{crassidis1996attitude}
J.~L. Crassidis and F.~L. Markley, ``Attitude estimation using modified rodrigues parameters,'' in \emph{Flight Mechanics/Estimation Theory Symposium 1996}, 1996.

\bibitem{makoviychuk2021isaac}
V.~Makoviychuk, L.~Wawrzyniak, Y.~Guo, M.~Lu, K.~Storey, M.~Macklin, D.~Hoeller, N.~Rudin, A.~Allshire, A.~Handa \emph{et~al.}, ``Isaac gym: High performance gpu-based physics simulation for robot learning,'' \emph{arXiv preprint arXiv:2108.10470}, 2021.

\bibitem{song2020denoising}
J.~Song, C.~Meng, and S.~Ermon, ``Denoising diffusion implicit models,'' \emph{arXiv preprint arXiv:2010.02502}, 2020.

\bibitem{kirillov2023segment}
A.~Kirillov, E.~Mintun, N.~Ravi, H.~Mao, C.~Rolland, L.~Gustafson, T.~Xiao, S.~Whitehead, A.~C. Berg, W.-Y. Lo \emph{et~al.}, ``Segment anything,'' \emph{arXiv preprint arXiv:2304.02643}, 2023.

\end{thebibliography}
